\documentclass[sigconf]{acmart}

\usepackage{booktabs} % For formal tables
\usepackage{url}  %Required
\usepackage{graphicx}  %Required
\usepackage{latexsym}
\usepackage{booktabs} % For formal tables
\usepackage{amsmath}
\usepackage{multirow}
\usepackage{braket}
\usepackage{amssymb}
\usepackage{bm}
\usepackage{color}
\usepackage{appendix} % Appendices

\setcopyright{rightsretained}
\begin{document}
\copyrightyear{2018} 
\acmYear{2018} 
\setcopyright{acmlicensed}
\acmConference[CIKM '18]{The 27th ACM International Conference on Information and Knowledge Management}{October 22--26, 2018}{Torino, Italy}
\acmBooktitle{The 27th ACM International Conference on Information and Knowledge Management (CIKM '18), October 22--26, 2018, Torino, Italy}
\acmPrice{15.00}
\acmDOI{10.1145/3269206.3271723}
\acmISBN{978-1-4503-6014-2/18/10}

\title{A Quantum Many-body Wave Function Inspired  \\ Language Modeling Approach}

\author{Peng Zhang$^{1}$, Zhan Su$^{1}$, Lipeng Zhang$^{2}$, Benyou Wang$^3$, Dawei Song$^{4}$ }
\authornote{Corresponding authors: P. Zhang and D. Song.}
\affiliation{
  \institution{$^{1}$ School of Computer Science and Technology, Tianjin University, Tianjin, China}
  \institution{$^{2}$ School of Computer Software, Tianjin University, Tianjin, China}
  \institution{$^{3}$ Department of Information Engineering, University of Padova, Padova, Italy}
  \institution{$^{4}$ School of Computer Science and Technology, Beijing Institute of Technology, Beijing, China}
}
\email{{pzhang,suzhan,lpzhang}@tju.edu.cn}
\email{wang@dei.unipd.it; dawei.song2010@gmail.com}

% \author{Peng Zhang}
% \authornote{Corressonding author.}
% \affiliation{%
%  \institution{School of Computer Science and Technology, Tianjin University, Tianjin, China}
% }
% \email{pzhang@tju.edu.cn}

% \author{Zhan Su}

% \affiliation{%
%   \institution{School of Computer Science and Technology, Tianjin University, Tianjin, China}
% }
% \email{suzhan@tju.edu.cn}

% \author{Lipeng Zhang}
% \affiliation{%
%   \institution{School of Computer Software, Tianjin University, Tianjin, China}
% }
% \email{lpzhang@tju.edu.cn}

% \author{Bengyou Wang}
% \affiliation{
%   \institution{School of Department of Information Engineering, University of Padova}
% }
% \email{wang@dei.unipd.it}

% \author{Dawei Song}
% \authornote{Corressonding author.}
% \affiliation{
%   \institution{School of Computer Science and Technology, Beijing Institute of Technology, China}
% }
% \email{dawei.song2010@gmail.com}

\begin{abstract}

The recently proposed quantum language model (QLM) aimed at a principled approach to modeling term dependency by applying the quantum probability theory. The latest development for a more effective QLM  has adopted word embeddings as a kind of global dependency information and integrated the quantum-inspired idea in a neural network architecture. While these quantum-inspired LMs are theoretically more general and also practically effective, they have two major limitations. First, they have not taken into account the interaction among words with multiple meanings, which is common and important in understanding natural language text. Second, the integration of the quantum-inspired LM with the neural network was mainly for effective training of parameters, yet lacking a theoretical foundation accounting for such integration. To address these two issues, in this paper, we propose a Quantum Many-body Wave Function (QMWF) inspired language modeling approach. The QMWF inspired LM can adopt the tensor product to model the aforesaid interaction among words. It also enables us to reveal the inherent necessity of using Convolutional Neural Network (CNN) in QMWF language modeling. Furthermore, our approach delivers a simple algorithm to represent and match text/sentence pairs. Systematic evaluation shows the effectiveness of the proposed QMWF-LM algorithm, in comparison with the state of the art quantum-inspired LMs and a couple of CNN-based methods, on three typical Question Answering (QA) datasets.

\end{abstract}

%It is, indeed, a challenging problem to bridge the quantum-inspired idea, the language modeling, and the neural network, in a principled manner.
%
% The code below should be generated by the tool at
% http://dl.acm.org/ccs.cfm
% Please copy and paste the code instead of the example below.
%
%\begin{CCSXML}
%<ccs2012>
% <concept>
%  <concept_id>10010520.10010553.10010562</concept_id>
%  <concept_desc>Computer systems organization~Embedded systems</concept_desc>
%  <concept_significance>500</concept_significance>
% </concept>
% <concept>
%  <concept_id>10010520.10010575.10010755</concept_id>
%  <concept_desc>Computer systems organization~Redundancy</concept_desc>
%  <concept_significance>300</concept_significance>
% </concept>
% <concept>
%  <concept_id>10010520.10010553.10010554</concept_id>
%  <concept_desc>Computer systems organization~Robotics</concept_desc>
%  <concept_significance>100</concept_significance>
% </concept>
% <concept>
%  <concept_id>10003033.10003083.10003095</concept_id>
%  <concept_desc>Networks~Network reliability</concept_desc>
%  <concept_significance>100</concept_significance>
% </concept>
%</ccs2012>
%\end{CCSXML}
%
%\ccsdesc[500]{Computer systems organization~Embedded systems}
%\ccsdesc[300]{Computer systems organization~Redundancy}
%\ccsdesc{Computer systems organization~Robotics}
%\ccsdesc[100]{Networks~Network reliability}
%

\keywords{Language modeling, quantum many-body wave function, convolutional neural network}

\maketitle

\section{Introduction}

It is essential to model and represent a sequence of words for many Information Retrieval (IR) or Natural Language Processing (NLP) tasks. In general, Language Modeling (LM) approaches utilize probabilistic models to measure the \textit{uncertainty} of a text (e.g., a document, a sentence, or some keywords). Based on different probability measures, there are roughly two different categories of LM approaches, namely traditional LMs~\cite{DBLP:series/synthesis/2008Zhai} based on the classical probability theory, and quantum-inspired LMs~\cite{sordoni2013modeling,aaaiZhang18} motivated by the quantum probability theory, which can be considered as a generalization of the classical one~\cite{Melucci2011,sordoni2013looking}.

% Essentially, the latter (based on subspaces) can be considered as a generalization of the former (based on subsets)~\cite{Melucci2011,sordoni2013looking}.
%

%Among those approaches, statistical language modelingis widely used approach to modeling such uncertainties by computing the probabilities of single words or compound words~\cite{DBLP:series/synthesis/2008Zhai}.

Recently, Sordoni, Nie and Bengio proposed a Quantum Language Modeling (QLM) approach, which aims to model the term dependency in a more principled manner~\cite{sordoni2013modeling}. In traditional LMs, modeling word dependency will increase the number of parameters to be estimated for compound dependencies (e.g., $n$-gram LM for IR)~\cite{DBLP:conf/sigir/SongC99}), or involve computing additional scores from matching compound dependencies in the final ranking function (e.g., Markov Random Field  based LM~\cite{DBLP:conf/sigir/MetzlerC05}). To solve these problems, QLM estimates a density matrix, which has a fixed dimensionality and encodes the probability measurement for both single words and compound words. In addition to its theoretical benefits, QLM has been applied to ad-hoc information retrieval task and achieved effective performance.

%In the IR literature, one fundamental research topic is to integrate the semantic dependency information in the language modeling approaches~\cite{DBLP:conf/sigir/SongC99,DBLP:conf/sigir/MetzlerC05,sordoni2013modeling}. However, in traditional approaches, modeling word dependency will increase the number of parameters to be estimated for compound dependencies (e.g., $n$-gram LM for IR)~\cite{DBLP:conf/sigir/SongC99}), or involve computing additional scores from matching compound dependencies in the final ranking function (e.g., Markov Random Field  based LM~\cite{DBLP:conf/sigir/MetzlerC05}). In order to solve these problems, Sordoni, Nie and Bengio proposed a principled approach named Quantum Language Modeling (QLM)~\cite{sordoni2013modeling}. QLM estimates a density matrix, which has a fixed dimensionality and encodes the probability measurement for both single words and compound words. However, in QLM, each single word is represented as an one-hot vector. The compound words are selected on two or more single words co-occurring in a fixed window size~\cite{sordoni2013modeling}, which essentially represent the local dependence (i.e., local co-occurrence), rather than a global semantic dependency between words.
%
In order to further improve the practicality of the quantum language models, a Neural Network based Quantum-like Language Model (NNQLM) was proposed~\cite{aaaiZhang18}. NNQLM utilizes word embedding vectors~\cite{mikolov2013distributed} as the state vectors, based on which a density matrix can be directly derived and integrated into an end-to-end Neural Network (NN) structure. NNQLM has been effectively applied in a Question Answering (QA) task. In NNQLM, a joint representation based on the density matrices can encode the similarity information of each question-answer pair. A Convolutional Neural Network (CNN) architecture is adopted to extract useful similarity features from such a joint representation and shows a significant improvement over the original QLM on the QA task.

%The distributional word representation, e.g., word embeddings, is considered as a richer representation capturing global semantic dependency~\cite{mikolov2013distributed}. More recently, 

Despite the progress in the quantum-inspired LMs from both theoretical and practical perspectives, there are still two major limitations, in terms of the representation capacity and seamless integration with neural networks. First, both QLM and NNQLM have not modeled the \textit{complex interaction} among words with multiple meanings. For example, suppose we have two polysemous words A and B, in the sense that A has two meanings $A_1$ and $A_2$, while B has two meanings $B_1$ and $B_2$. If we put them together and form a compound word, this compound word will have four possible states ($A_1B_1$, $A_1B_2$, $A_2B_1$, $A_2B_2$), each corresponding to a combination of specific meanings of different words. If we have more words, such an interaction will become more complex. However, in QLM and NNQLM, a compound word is modeled as a direct addition of the representation vectors or subspaces of the single words involved. Therefore, it is challenging to build a language modeling mechanism which has the representation capacity towards the complex interactions among words as described above.

Second, although in NNQLM the neural network structure can help quantum-inspired LM with effective training, the fundamental connection between the quantum-inspired LM and the neural network remains unclear. In other words, the integration of NN and QLM so far has not been in a principled manner. Hence, we need to investigate and explain the intrinsic rationality of neural network in quantum-inspired LM. It is challenging, yet important to bridge quantum-inspired idea, language modeling, and neural network structure together, and develop a novel LM approach with both theoretical soundness and practical effectiveness.

% In fact, the $explainability$ of neural network itself is still under-explored. 

In order to address the above two challenges, we propose a new language modeling framework inspired by Quantum Many-body Wave Function (QMWF). In quantum mechanics, the wave function can model the interaction among many spinful particles (or electrons), where each particle is laying on multiple states simultaneously, and each state corresponds to a basis vector~\cite{Science2017,Nielsen:2011}. Therefore, by considering a word as a particle, different meanings (or latent/embedded concepts) as different basis vectors, the interaction among words can be modeled by the tensor product of different basis vectors for different words. It is then natural to use such a QMWF formalism to represent the complex interaction system for a sequence of natural language words. 

%It is noting that each base vector can express an latent semantic space, which may be multiple meanings of a word, or it may be multiple topics in which a word is located. For example, \emph{Apple} can represent both fruits and technology companies. It is in the semantic space of fruits and technology companies spanned by their basic vector respectively.

In addition, we show that the convolutional neural network architecture can be mathematically derived in our quantum-inspired language modeling approach. Since the tensor product is performed in QMWF based LM, the dimensionality of the tensor will be  exponentially increased, yielding a quantum many-body problem. To solve this problem, the tensor decomposition can be used to solve a high-dimensional tensor~\cite{LevineYCS17}. With the help of tensor decomposition, the projection of the global representation to the local representation of a word sequence can result in a 
convolutional neural network architecture. In turn, for the convolutional neural network, it can be interpreted as a mapping form a global semantic space to the local semantic space. 

%To address the exponentially increasing dimensionality when the tensor product is performed, the tensor decomposition yields an approximate solution (based on the convolutional neural network) for the quantum many-body problem~\cite{Science2017,LevineYCS17}. Thus, the proposed framework will provide a fundamental connection between QMWF based language modeling and the neural network architecture.

%The proposed language modeling approach, to some extent, can also include the classical n-gram language modeling as its special case\pz{still needs to be confirmed}. Practically,

Hence, our QMWF inspired Language Modeling (QMWF-LM) approach also delivers a feasible and simple algorithm to represent a text or a sentence and match the text/sentence pairs, in both word-lever and character-level. We implement our approach in the Question Answering task. The experiments have shown that the proposed QMWF-LM can not only outperform its quantum LM counterparts (i.e., QLM and NNQLM), but also achieve  better performance in comparison with typical CNN-based approaches on three QA datasets.

Our main contributions can be summarized as follows:
\begin{enumerate}
  \item We propose a Quantum Many-body Wave Function based Language Modeling (QMWF-LM) approach, which is able to represent complex interaction among words, each with multiple semantic basis vectors (for multiple meanings/concepts).
  \item We show a fundamental connection between QMWF based language modeling approach and the convolutional neural network architecture, in terms of the projection between the global representation to the local one of a word sequence. 
  \item The proposed QMWF-LM delivers an efficient algorithm to represent and match the text/sentence pairs, in both word-lever and character-level, as well as achieves effective performance on a number of QA datasets.
\end{enumerate}

%and bridging the quantum-inspired idea, the language modeling and the neural network designing

%Since the quantum mechanics can provide more insights and explainability, for example, the QMB with the Recurrent neural network, we believe that this work is the first attempt, and will

%In the language modeling scenario, we show that the approximated solution for the QMB wave function (projected on its product state) share the similar ideas with the convolutional neural network (CNN) based methods in natural language processing tasks, e.g., Question Answering tasks~\cite{hu2014convolutional,kim2014convolutional,yin2015abcnn}. In turn, such an observation, can entail the CNN based language modeling with more physical meanings and explainability.
%
%
%\textbf{Clarify the difference between QMB-LM with QLM and NNQLM}
%
%\textbf{Make clear the contribution of our proposed work on LM}
%
%\textbf{Add a figure about the relations among LM, QT, NN, make clear our contributions}
%

%Since the quantum mechanics can provide more insights and explainability, for example, the QMB with the Recurrent neural network, we believe that this work is the first attempt, and will open a door for the consequent study on the cross fields among the language modeling, neural network, and quantum mechanics.

\section{Quantum Preliminaries}
In this section, we first describe some basis notations and concepts of the quantum probability theory. Then, we briefly explain the quantum many-body wave function.

\subsection{Basic Notations and Concepts} \label{section:basic notation}

%Now, we show the Dirac notations and descriptions for the state vector, superposition, pure states, mixed states, as well as the corresponding probability measurements.

The formalism of quantum theory is actually based on vector spaces using Dirac notations. In line with previous studies on the quantum-inspired language models~\cite{sordoni2013modeling,sordoni2014learning,aaaiZhang18}, we restrict our problem to vectors spaces over real numbers in $\mathbb{R}$.

A wave function in quantum theory is a mathematical description of the quantum state of a system. A state vector
is denoted by a unit vector $\ket{\psi}$ (called as a $ket$), which can be considered as a column vector $\vec{\psi} \in \mathbb{R}^n$ (for better understanding). The transpose of $\ket{\psi}$ is denoted as $\bra{\psi}$ (called as $bra$), which is a row vector.

The state vector can be considered as a $ray$ in a Hilbert space (i.e., $\ket{\psi} \in  \mathcal{H}$), which has a set of orthonormal basis vectors $\ket{e_i} (i = 1,\ldots,n)$. A state vector $\ket{\psi}$ can be a superposition of the basis vectors:
\begin{equation}
\label{eq:superposition}
\ket{\psi} = \sum_{i=1}^n a_i \ket{e_i}
\end{equation}
where $a_i$ is a probability amplitude and $\sum_{i} a_i ^2 = 1$, since $a_i ^2$ represents a probability of the sum to 1. 

For example, suppose we have a two basis vectors $\ket{0}$ and $\ket{1}$, which can be considered as $(1,0)^T$ and $(0,1)^T$, respectively. Then, we have a state vector
 \begin{displaymath}
 \ket{\psi} = a_1 \ket{0} + a_2 \ket{1}
 \end{displaymath}
 It means that the corresponding quantum system $\ket{\psi}$ is a \textit{superposed state}, i.e., it is in the two states $\ket{0}$ and $\ket{1}$ simultaneously. In the natural language processing tasks, such a superposed state can be used to model the multiple semantic meanings of a word~\cite{blacoe2013quantum}. 

% The projection can be formulated by the \textit{inner product} in the Hilbert space. 
 
 %Specifically, 
 
 \begin{figure}[t]
\centering
\includegraphics[scale=0.5]{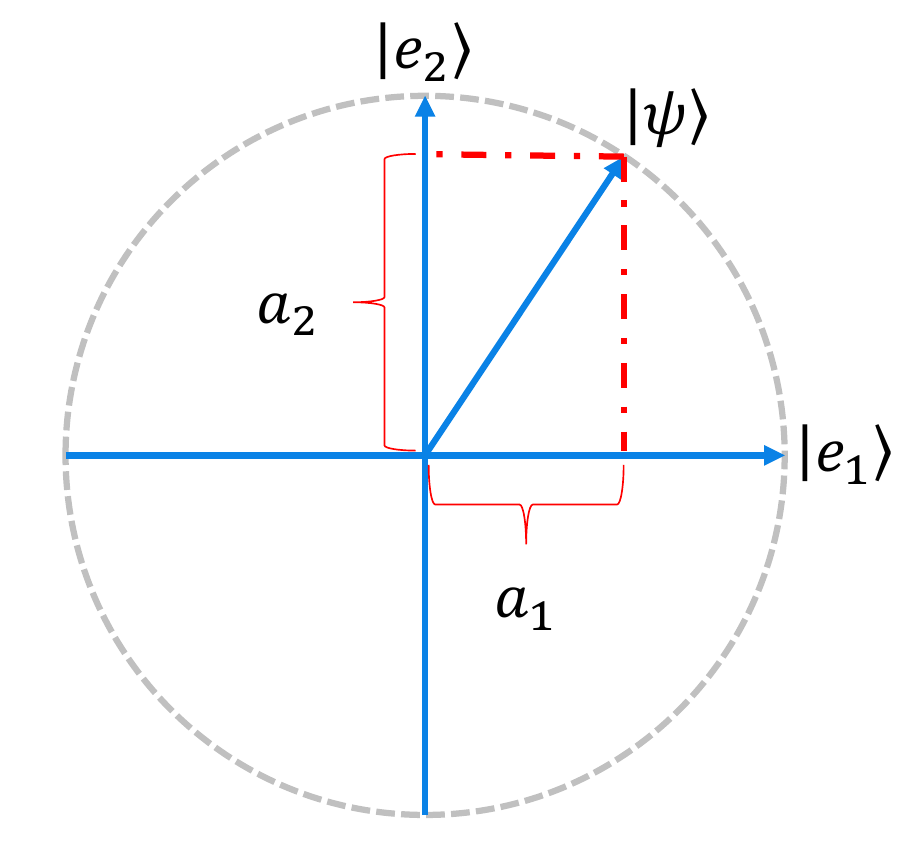}
\caption{Projection of $\ket{\psi}$ on its basis}
\label{fig:figure-projection}
\end{figure}

 In Eq.~\ref{eq:superposition}, the  probability amplitude $a_i$ can be calculated by the \textit{inner product} $\braket{e_i|\psi}$.
\begin{displaymath}
 a_i  = \braket{e_i|\psi}
\end{displaymath}
The inner product $\braket{e_i|\psi}$ is a projection of $\ket{\psi}$ onto $\ket{e_i}$. As illustrated in Fig.\ref{fig:figure-projection}, the projection measurement can be formulated as 
\begin{equation}
\label{eq:projection_measure}
p(e_i|\psi) = a_i^2 = \braket{e_i|\psi}^2
\end{equation}
where $p(e_i|\psi)$ denotes the probability of the quantum elementary event $\ket{e_i}$ \footnote{More strictly, the outer product of $\ket{e_i}$ is called as the quantum elementary event.} given the system $\ket{\psi}$. 

It turns out that the projection measurement based on inner product plays an essential role in the probability measurement, We will further illustrate such a concept in our quantum many-body wave function inspired LM approach. Note that, in a broad sense, the \textit{wave function} is a state vector $\ket{\psi}$. In a narrow sense, the wave function is a projection on a basis, e.g., $\psi(x) = \braket{x|\psi}$, where $x$ can be a basis $e_i$, and $\psi(x)$ is the probability amplitude. In this paper, we will use the description of wave function in the broad sense.

%According to the previous discussion about the projection measurement, $a_1$ can be obtained by the inner product $\braket{0|\psi}$.

%Generally, \textit{quantum superposition} allows the possibility of the multiple exclusive states simultaneously in a single state~\cite{Nielsen:2011}. 

% \begin{comment}
% In addition, we can define an outer product $\ket{e_i}\!\bra{e_i}$, which is also called as a dyad $\mathit{\bm{\Pi}_i}$ of the basis $\ket{e_i}$. Such a dyad is corresponding to a quantum elementary event~\cite{sordoni2013modeling}, and is also projector. To see this, we have
% \begin{equation}
% \label{eq:projection_measure}
% p(e_i|\psi) = a_i^2 = \| \mathit{\bm{\Pi}_i}\ket{\psi}\|_2^2 = \braket{e_i|\psi}^2
% \end{equation}

% \end{comment}

% %\begin{displaymath}
% %\mathit{\bm{\Pi}_i}\ket{\psi} = \ket{e_i}\!\bra{e_i} \ket{\psi} =  \ket{e_i}\!\braket{e_i|\psi} = a_i \ket{e_i}
% %\end{displaymath}
% %Such a projection is illustrated in fig.\ref{fig:figure-projection}. In quantum theory, we have

% %The ranking is based on the matching score, which is computed by a neural network structure built on the joint representation of the question and the answer.

% %
% %The density matrix in Eq.\ref{eq:NNQLMdensity} can be derived analytically, thus making feasible for an integration in an End-to-End design based on Neural Network.

\subsection{Quantum Many-Body Wave Functions}
What we mentioned above is a single system which corresponds to a single particle in a Hilbert space. A quantum many-body system consists of $N$ particles, each one with a wave function residing in a finite dimensional Hilbert space $\mathcal{H}_i$ for $i\in [N]:=\{1\ldots N\}$. We set the dimensions of each Hilbert space $\mathcal{H}_i$ for all $i$, i.e., $\forall i$ : $dim(\mathcal{H}_i)=M$ and the orthonormal basis of the Hilbert space as $\{|e_h\rangle\}_{h=1}^{M}$. The Hilbert space of a many-body system is a \textit{tensor product}  of the spaces: $\mathcal{H}:=\otimes_{i=1}^{N}\mathcal{H}_i$, and the corresponding state vector $\ket{\psi}\in\mathcal{H}$ is
\begin{equation}
\label{eq:QMWF}
\ket{\psi}=\sum_{h_1,\ldots,h_N=1}^{M}\mathcal{A}_{h_1\ldots h_N}\ket{e_{h_1}}\otimes\cdots\otimes\ket{e_{h_N}}
\end{equation}
where $|e_{h_1}\rangle\otimes\cdots\otimes|e_{h_N}\rangle$ is a basis vector of the $M^N$ dimensional Hilbert space $\mathcal{H}$, and $\mathcal{A}_{h_1\ldots h_N}$ is a specific entry in a tensor $\mathcal{A}$ holding all the probability amplitude. The tensor $\mathcal{A}$ can be considered as $N$-dimensional array $\mathcal{A}\in{\mathbb{R}^{M\times\cdots\times M}}$.

For example, a system includes two spinful particles, which are qubit states superposed as two basis vectors $\ket{e_1}=\ket{0}={(1,0)}^T$ and $\ket{e_2}=\ket{1}={(0,1)}^T$. Therefore, we can get four basis vectors $\ket{e_1}\otimes\ket{e_1}=\ket{00}$ (abbreviation of $\ket{0}\otimes\ket{0} = {(1,0,0,0)}^T$), $\ket{01} = {(0,1,0,0)}^T$, $\ket{10}= {(0,0,1,0)}^T$ and $\ket{11}= {(0,0,0,1)}^T$, and the state $\psi$ of this system can be represented as 
\begin{equation}
\begin{aligned}
\label{eq:QMWF-simple}
\ket{\psi} &= \sum_{i,j=1}^{2} a_{ij} \ket{e_i}\otimes\ket{e_j} \\
&= a_{11}\ket{00} + a_{12}\ket{01} + a_{21}\ket{10} + a_{22}\ket{11}
\end{aligned}
\end{equation}
where $a_{ij}$ is a probability amplitude and $\sum_{ij} a_{ij} ^2 = 1$. Each $a_{ij}$ can be considered as a specific entry in a tensor $\mathcal{A}\in\mathbb{R}^{2\times 2}$. 

%Consequently, if a system has three particles, we have $2^3=8$ basis vectors.

%Furthermore, suppose a quantum system has $N$ particles, and set the dimensions of each Hilbert space $\mathcal{H}_i$ for all $i$, i.e., $\forall i$ : $dim(\mathcal{H}_i)=M$. One can get the $M^N$ basis vectors for the system and a probability amplitude tensor $\mathcal{A}\in{\mathbb{R}^{M\times\cdots\times M}}$. %The more details of the formula symbol are shown in the appendix.

%We have described a quantum many-body wave function to represent a system containing $N$ particles. It inspired us to model the complex interaction between words with QMWF on account of the tensor product of different basis vectors, which represent multiple sementic meanings.

%for the system. Then, we have
%\begin{displaymath}
%\ket{e_1}\otimes\ket{e_1}=\ket{0}\otimes\ket{0}=\begin{bmatrix} 1 \\ 0 \end{bmatrix} \otimes \begin{bmatrix} 1 \\ 0 \end{bmatrix} = \begin{bmatrix} 1\times1 \\ 1\times0 \\ 0\times1 \\ 0\times0 \end{bmatrix} = \begin{bmatrix} 1 \\ 0 \\ 0 \\ 0 \end{bmatrix}
%\end{displaymath}
%and a 

\section{Quantum Many-Body Wave Function Inspired Language Modeling}
\label{section:QMWF-framework}

\subsection{Basic Intuitions and Architecture}
\label{sec:basic_intuitions}

In Physics, Quantum Many-body Wave Function (QMWF) can model the interaction among many particles and the associated basis vectors. In the language scenario, by considering a word as a particle, different meanings (or latent/embedded concepts) as different basis vectors, the \textbf{interaction} among words (or word meanings) can be modeled by the \textbf{tensor product} of basis vectors, via the many-body wave function. A tensor product of different basis vectors generates \textit{a compound meaning} for a compound word. 

Based on such an analogy, \textbf{QMWF representation} can model the \textbf{probability distribution} of \textbf{compound meanings} in natural language. Such a representation depends on \textbf{basis vectors}. The choices of basis vectors can be one-hot vectors (representing single words), or embedding vectors (representing latent meanings or concepts). The \textbf{probabilities} are encoded in a \textbf{tensor}, as we can see from Eq.~\ref{eq:QMWF}. Each entry in a tensor is the probability amplitude of the compound meaning, or can be considered as a coefficient/weight.

\begin{figure*}[htp]
\centering
\includegraphics[scale=0.5]{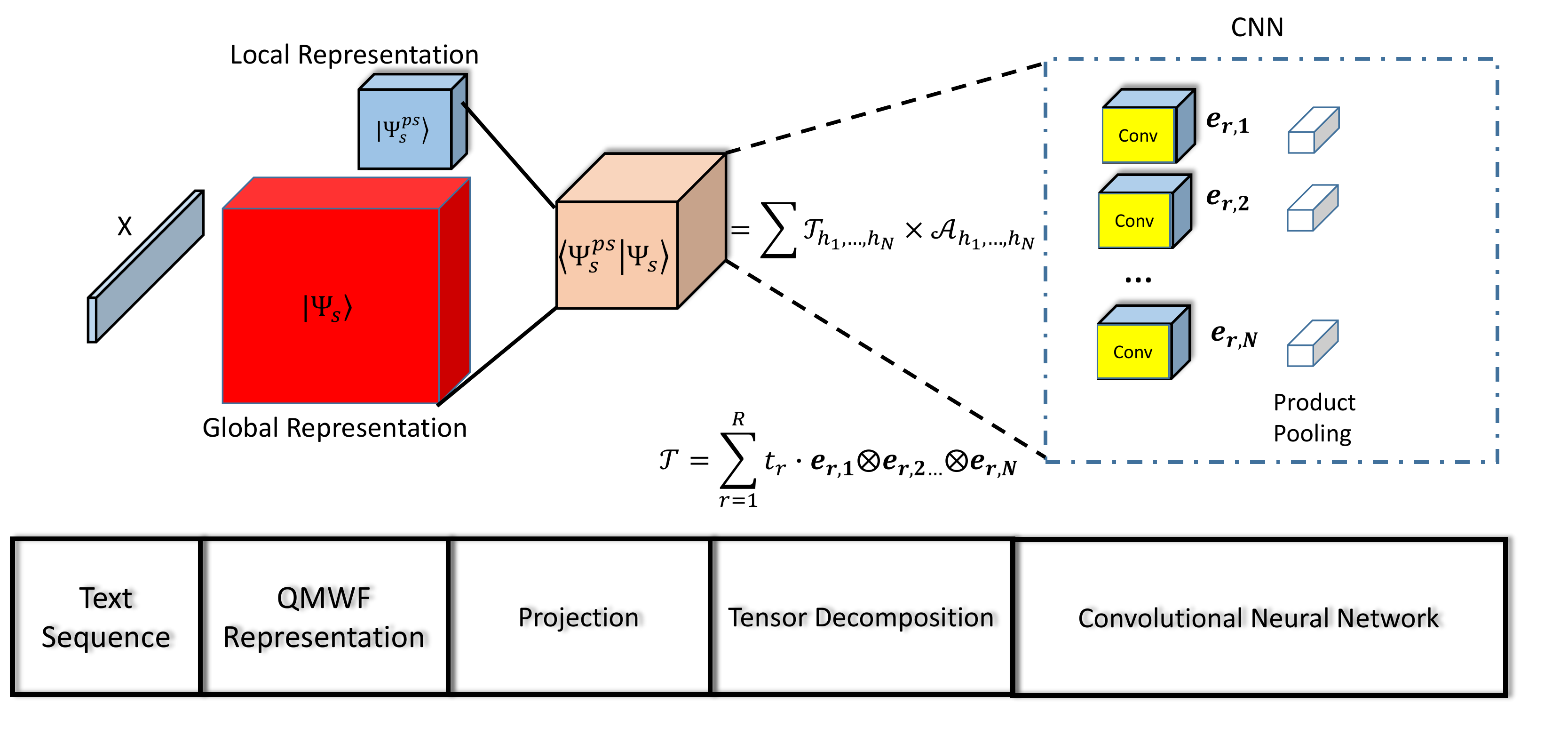}
\caption{Outline of quantum many-body wave function inspired language modeling approach}
% \caption{ (a) It shows the projection of $\ket{\psi_S}$ onto $\ket{\psi_S^{ps}}$, and the each step can be considered as a layer in neural network architecture. (b) After decomposing the weight tensor $\mathcal{T}$, it reveals a convolutional network architecture.}
\label{fig:figure-all}
\end{figure*}

As shown in Fig. \ref{fig:figure-all}, given a word sequence and a set of basis vectors, there are \textbf{local} and \textbf{global representations} (see details in Section~\ref{section: text2QLM}). Intuitively, the local representation is constructed by the current word sequence (e.g., a sentence), and the global representation corresponds to the information of a large corpora (e.g., a collection of data). In classical language modeling approaches, there are also local and global information, as well as the interplay between them. For example, in  $n$-grams, the probability/statistics of each term can be estimated from the current piece of text (as local information), and also be smoothed with the statistics from a large corpora (as global information). 

%The smoothing can be considered as a kind of interplay between the local information and the global one. 

Based on QMWF representation, in Section~\ref{section:product-state}, we describe the \textbf{projection} from the global representation to the local one. Such projection can model the interplay between the local information and the global one, and enable us to focus on the high-dimensional tensors, which 
encode the probability distribution of the compound meanings. In Fig. \ref{fig:figure-all}, we can observe that, the high-dimensional tensor $\mathcal{T}$ can be reduced by the tensor decomposition, the tensors $\mathcal{A}$ (for probabilities in local representation) and $\mathcal{T}$ (for probabilities in global representation) are kept. Therefore, the projection can also be considered as an interplay between global and local tensors (see Section~\ref{section:product-state} for details).

% <<<<<<< HEAD
% After the projection, the high-dimensional tensor remained as a many-body problem to be solved. It is shown that \textbf{tensor decomposition} can be adopted to approximate the tensor. 
% With the help of the tensor decomposition, the \textbf{convolutional neural network} architecture can be realized as a projection (or a mapping intuitively) from the global representation (as a global semantic space) to the local one. (see Section~\ref{section:ten-dec-con}). For example, the subspaces in tensor decomposition corresponds to the kernels of the convolutional layer. The derivation of the projection also yields a product pooling layer. Then, an algorithm based on a CNN architecture is revealed based on the above intuitions.
% =======
The high-dimensional tensor $\mathcal{T}$ can be reduced by the \textbf{tensor decomposition}.  
With the help of the tensor decomposition, the above projection  from the global representation (as a global semantic space) to the local one can be realized by a \textbf{convolutional neural network} architecture(see Section~\ref{section:ten-dec-con}). Intuitively, each decomposed subspace of the high-dimensional tensor corresponds to a convolutional channel. Together with a product pooling technique, a CNN architecture can be constructed. Then, an algorithm based on a CNN architecture is revealed based on the above intuitions.

\subsection{Language Representation and Projection via Many-body Wave Function} \label{section: text2QLM}

\subsubsection{Local Representation by Product State}

Suppose we have a word sequence $S$ (e.g., a sentence) with the length $N$: $S = [x_{1},x_{2},...,x_{N}]$. For each word $x_i$ in $S$, based on Eq.~\ref{eq:superposition}, we define its state vector as:
\begin{equation}
\label{eq:singlewave_index}
\ket{x_i}=\sum_{h_i=1}^{M}\alpha_{i,h_i}\ket{\phi_{h_i}}
\end{equation}
where each basis vector $\ket{\phi_{h_i}} (h_i=1,\ldots,M)$ 
 is corresponding to a specific semantic meaning (or a latent concept), and $\alpha_{i,h_i}$ is its associated probability amplitude. Different from the notation $\alpha_i$ in Eq.~\ref{eq:superposition}, the notation $\alpha_{i,h_i}$  in Eq.\ref{eq:singlewave_index} is for the convenience to be represented in a tensor depicted latter. 
For a better understanding, as an example, the state vectors for words $x_1$ and $x_2$ can be represented by
\begin{equation}
\label{eq:exampleforwave}
\left\{
             \begin{array}{lr}
             \ket{x_1}=\alpha_{1,1}\ket{\phi_1}+\alpha_{1,2}\ket{\phi_2}+\alpha_{1,3}\ket{\phi_3} &  \\
             \ket{x_2}=\alpha_{2,1}\ket{\phi_1}+\alpha_{2,2}\ket{\phi_2}+\alpha_{2,3}\ket{\phi_3} &
             \end{array}
\right.
\end{equation}
where $N=2$,  $M=3$ and $h_i$ is from 1 to 3, i.e., three basis vectors $\ket{\phi_{h_i}}$ are involved, each corresponding to a word meaning.

For the basis vectors $\ket{\phi_{h_i}}$, there can be different choices, e.g., one-hot vectors or embedded vectors. Different basis vectors yield different interpretations for the semantic meanings. We will adopt the embedding space when we instantiate this framework in the question answering task (see Section~\ref{sec:appQA}). If we use such a space, the probability amplitude $\alpha_{i,h_i}$ is the feature value (after normalization) of the word $x_i$ on the ${h_i}^{th}$ dimension of the embedding space.

%(e.g., $\alpha_{1,3}$ for $\ket{\phi_3}$)

%Note that some of the above amplitudes $\alpha_{i,h_i}$ can be zero, since a specific word may not contain all the possible meanings/concepts.

%By considering a sentence is a many-body system and each word is a single system, in quantum mechanics, the product state means that  $\ket{\psi_S^{ps}}$ is not entanglement state since it can be decomposed into several subsystem $\ket{x_i}$. In this paper, we will not discuss the entanglement. Our focus is on the wave function representation and the calculation of the concerned probability amplitudes.

Next, we show how to use the tensor product to model the interaction among word meanings. For a sentence $S = [x_{1},x_{2},...,x_{N}]$, its wave function can be represented as:
\begin{equation}
\label{eq:ps_state}
\ket{\psi_S^{ps}}=|x_{1}\rangle\otimes\ldots\otimes|x_{N}\rangle
\end{equation}
where $\ket{\psi_S^{ps}}$ is the \textit{product state} of the QMWF representation of a sentence. We can expand the product state $\ket{\psi_S^{ps}}$ as follows:
\begin{equation}
\label{eq:text-rep-ps}
\ket{\psi_S^{ps}} = \sum_{h_1,\ldots,h_N=1}^{M}\mathcal{A}_{h_1\ldots h_N}|\phi_{h_1}\rangle\otimes\ldots\otimes|\phi_{h_N}\rangle
\end{equation}
where $\ket{\phi_{h_1}}\otimes\ldots\otimes \ket{\phi_{h_N}}$ is the new basis vectors with $M^N$ dimension, and each new basis vector corresponds a \textit{compound meaning} by the tensor product of the word meanings  $\ket{\phi_{h_i}}$.
$\mathcal{A}$ is a $M^N$ dimensional tensor and each entry ${\mathcal{A}_{h_1\ldots h_N}}$ ($= \prod_{i=1}^{N}\alpha_{i,h_i}$) encodes the probability of the corresponding compound meaning.

%the corresponding probability amplitude of each basis vector. We have
%\begin{displaymath}
%\mathcal{A}_{h_1\ldots h_N} = \prod_{i=1}^{N}\alpha_{i,h_i}
%\end{displaymath}
%which encodes the probability of the corresponding compound meaning. 

Eq.~\ref{eq:text-rep-ps} can  represent the interaction among words as we discussed in Introduction. For example, for two words $x_1$ and $x_2$ in Eq.~\ref{eq:exampleforwave}, suppose $x_1$ only has two meanings corresponding to the basis vectors $\ket{\phi_1}$ and $\ket{\phi_2}$, while  $x_2$ has two meanings corresponding to $\ket{\phi_2}$ and $\ket{\phi_3}$. Then, $\mathcal{A}_{1,3} (=\alpha_{1,1} \alpha_{2,3})$ represents the probability with the basis vector $\ket{\phi_1}\otimes\ket{\phi_3}$. Intuitively, this implies that the underlying meaning ($\ket{\phi_1}$) of word  $x_1$ and the meaning ($\ket{\phi_3}$) of $x_2$ is interacted and form a \textit{compound meaning} $\ket{\phi_1}\otimes\ket{\phi_3}$.

Now, we can see that this product state representation is actually a local representation for a word sequence. In other words, given the basis vectors, the probability amplitudes can be estimated from the current word sequence. In fact, 
$\mathcal{A}$ is a \textit{rank-1} $N$-order tensor, which includes only $M\times N$ free parameters $\alpha_{i, h_i}$, rather than $M^N$ parameters to be estimated.  In addition, given only a word sequence, the valid compound meanings are not too many. In summary, this rank-1 tensor actually encodes the local distributions of these compound meanings for the given sentence.

%For sentences $S$ with length of $N$, $S = [x_{1},x_{2},...x_{n}]$, each word has a M semantic space representation. The semantic space of whole sentence is $M^N$.

\begin{comment}
As shown in figure \ref{fig:figure-path}, words will reside in one of the M semantic spaces with a certain probability. The same as other words in sentence. A path, as shown in figure \ref{fig:figure-path}, represents such a semantic space combination with interactive information among words. In figure \ref{fig:figure-proj2conv}, the path is represented by $\prod\cdot$. Semantic information of whole sentence is reside in this semantic space with $M^N$ dimension.
\end{comment}

%\benyou{here  mention  $\braket{\psi_S^{ps} | \psi_S^{ps} }$  }

\subsubsection{Global Representation for All Possible Compound Meanings}
\label{sec:mbf}

As aforementioned in Section~\ref{sec:basic_intuitions} , we need a global distribution of all the possible compound meanings, given a set of basis vectors. Intuitively, a global distribution is useful in both classical LM and quantum-inspired LM, since we often have \textit{unseen} words, word meanings or the compound meanings, in a text.

% Then, if a new sentence comes, we can project the global Hilbert space to the local one and compute the probabilities $p_i$, which will be described in Section~\ref{section:product-state}.

To represent such a global distribution of state vectors, we define a quantum many-body wave function as follows:
\begin{equation}
\label{eq:text-rep}
\ket{\psi_S} = \sum_{h_1,\ldots,h_N=1}^{M}\mathcal{T}_{h_1\ldots h_N}|\phi_{h_1}\rangle\otimes\ldots\otimes|\phi_{h_N}\rangle
\end{equation}
where $\ket{\phi_{h_1}}\otimes\ldots\otimes \ket{\phi_{h_N}}$ is the basis state (corresponding to a compound meaning) with $M^N$ dimension, and ${\mathcal{T}_{h_1\ldots h_N}}$ is the corresponding probability amplitude. This wave function represents a semantic meaning space with a sequence of $N$ uncertain words, which does not rely on a specific sentence showed in Eq.~\ref{eq:text-rep-ps}. The probability amplitudes in tensor $\mathcal{T}$ can be trained in a large corpora. 

% with the probability as its square value, i.e., ${\mathcal{T}^2_{h_1\ldots h_N}}$.

%The entries of two tensors are probability amplitudes, which are the key components to be estimated in our approach. 

The difference between  $\ket{\psi_S^{ps}}$ in Eq.~\ref{eq:text-rep-ps} and $\ket{\psi_S}$ in Eq.~\ref{eq:text-rep} is the different tensors  $\mathcal{A}$ and $\mathcal{T}$.
$\mathcal{A}$  encodes the \textit{local} distribution of compound meanings (for the current sentence) and  $\mathcal{T}$ encodes the \textit{global} distribution (for a large corpora). Moreover, $\mathcal{A}$ is essentially rank-1, while $\mathcal{T}$ has a higher rank. In fact, solving $\mathcal{T}$ is an intractable problem  which is referred as a quantum many-body problem.

\subsubsection{Projection from Global to Local Representation} \label{section:product-state}

Section \ref{section:basic notation} has emphasized the role of projection in the probability measurement. Now, we show the projection of the global semantic representation $\ket{\psi_S}$ on its product state $\ket{\psi_S^{ps}}$ as a local representation for the given sentence, to calculate the probability amplitudes in the tensor.

Such a projection can be modeled by the inner product $\braket{\psi_S^{ps}|\psi_S}$. Inspired by a recent work~\cite{LevineYCS17}, this projection will eliminate the high-dimensional basis vectors of the wave function:
\begin{equation}
\label{eq:projection}
\begin{aligned}
\braket{\psi_S^{ps}|\psi_S}&=\bm{\langle} x_{1}\ldots x_{N}\bm{|}\sum_{h_1,\ldots,h_N=1}^M\mathcal{T}_{h_1\ldots h_N}\ket{\phi_{h_1}\ldots\phi_{h_N}}\bm{\rangle}\\
&= \sum_{h_1,\ldots,h_N=1}^{M}\mathcal{T}_{h_1\ldots h_N}\prod_{i=1}^{N}\braket{x_i|\phi_{h_i}}_i\\
&= \sum_{h_1,\ldots,h_N=1}^{M}\mathcal{T}_{h_1\ldots h_N}\prod_{i=1}^{N}\alpha_{i,h_i} \\ 
&= \sum_{h_1,\ldots,h_N=1}^{M}\mathcal{T}_{h_1\ldots h_N} \times \mathcal{A}_{h_1\ldots h_N} \\ 
\end{aligned}
\end{equation}
which reveals the interplay between the global tensor $\mathcal{T}$ and local tensor $\mathcal{A}$. This is similar to the idea in the classical LM, where the local statistics in a text will be smoothed by collection statistics.

\begin{comment}
Such a projection $\braket{\psi_S^{ps}|\psi_S}$ can be illustrated in Fig. \ref{fig:figure-all}. First, each word $\ket{x_i}$ is projected onto its basis vectors $\{\ket{\phi_h}\}_{h=1}^{M}$ to obtain a number ($N$) of $M$-dimensional vector $\bm{\alpha}_i=(\alpha_{i,1},\ldots,\alpha_{i,M})^T$ for $i\in[N]$. Second, one can choose an entry $\alpha_{i,h_i}$ from each vector $\bm{\alpha}_i$, and then multiply with each other, yielding a product $\alpha_{1,h_1}\cdots\alpha_{N,h_N}$, which is the entry $\mathcal{A}_{h_1,\ldots,h_N}$ of the tensor $\mathcal{A}$ in Eq.~\ref{eq:text-rep-ps} . Taking all the values $i$ from $1$ to $N$ and $h_i$ from $1$ to $M$, we will get a number ($M^N$) of products. Finally, these $M^N$ products are summed to get the final projection $\braket{\psi_S^{ps}|\psi_S}$ with the $M^N$-dimensional weight tensor $\mathcal{T}$. 
\end{comment}

%Although the high-dimensional bases have been eliminated, the high-dimensional tensor $\mathcal{T}$ is still an unsolved issue. Next, we will decompose the tensor to solve this problem.

\subsection{Projection Realized by Convolutional Neural Network} \label{section:ten-dec-con}

As shown in Eq. \ref{eq:projection}, the high-dimensional tensor $\mathcal{T}$ is still an unsolved issue. Now, we first describe the tensor decomposition to solve this high-dimensional tensor. Then, with the decomposed vectors, we will show that the convolutional neural network can be considered as a projection or a mapping process from the global semantics to local ones. 

\subsubsection{Tensor Decomposition} \label{Section:ten-dec}

 \begin{figure}[t]
 \centering
 \includegraphics[scale=0.35]{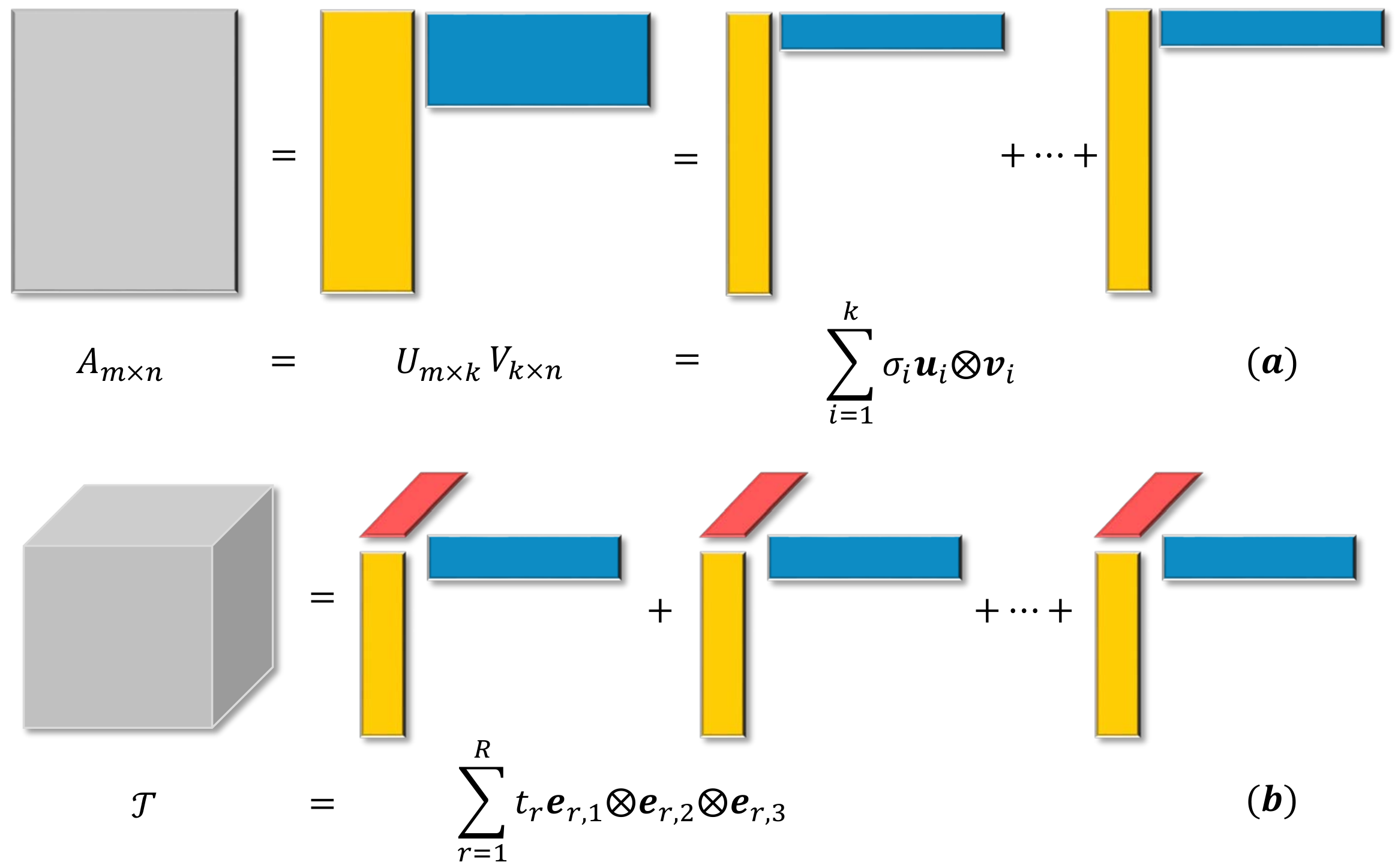}
 \caption{An illustration of the singular value decomposition of a matrix with dimension $M\times N$ in (a) and the CP decomposition of a three order tensor in (b).}
 \label{fig:figure-deco}
 \end{figure}
 
\begin{comment}
As shown in Eq. \ref{eq:projection}, $\braket{\psi_S^{ps}|\psi_S}$ is the projection of global semantics to local semantics space and tensor $\mathcal{T}$ has exponential entries, which are hardly computed. It is necessary to reduce the high-dimensional space and extracted the principle semantic information. 

In linear algebra, for matrix A, Let R be the rank of matrix A and its SVD is given by 
\begin{equation}
\label{eq:svd}
A = \sum^R_{r=1}\sigma_ru_r\otimes v_r, \quad  with \ \sigma_1 \geq \sigma_2 \geq ... \sigma_R > 0
\end{equation}
We can obtain a best rank-k approximation given by the leading k factors of the singular value decomposition (SVD) \cite{kolda2009tensor}. This is a process that can be used to reduce the high-dimensional space to a approximated low-rank space that still contains most of the information. Then a rank-k approximation is $A \approx \sum^k_{r=1}\sigma_ru_r\otimes v_r$.
\end{comment}

In general, \textit{Tensor decomposition} can be regarded as a generalization of Singular Value Decomposition (SVD) from matrices to tensors and can help to solve high-dimensional problems (see Fig.~\ref{fig:figure-deco}). There are many methods to decompose a high-dimensional tensor, such as Canonical Polyadic Decomposition (CP decomposition~\cite{hitchcock1927expression}), Tucker Decomposition, etc. The CP decomposition with weights is:
\begin{equation}
\label{eq:tensor-decomposition}
\mathcal{T}=\sum_{r=1}^{R}t_{r}\cdot\bm{e}_{r,1}\otimes\bm{e}_{r,2}\otimes\ldots\otimes\bm{e}_{r,N}
\end{equation}
where $t_r$ is the weight coefficient for each rank-1 tensor and $\bm{e}_{r,i}=(e_{r,i,1},\ldots,e_{r,i,M})^T$ ($i=1,\ldots,N$) is a unit vector with $M$-dimension. $R$ is the rank of $\mathcal{T}$, which is defined as the smallest number of rank-1 tensors that generate $\mathcal{T}$ as their sum.

The decomposed vector $\bm{e}_{r,i}$ with a low dimension will play a key role in the later derivation. A set of vectors $\bm{e}_{r,i}  (i=1,\ldots,N)$ can be a subspace of the high-dimensional tensor $\mathcal{T}$.

%, and the convolutional function can be considered as the mapping from a basis of such a subspace to the input representation of a word.

\begin{figure*}[t]
\centering
\includegraphics[scale=0.4]{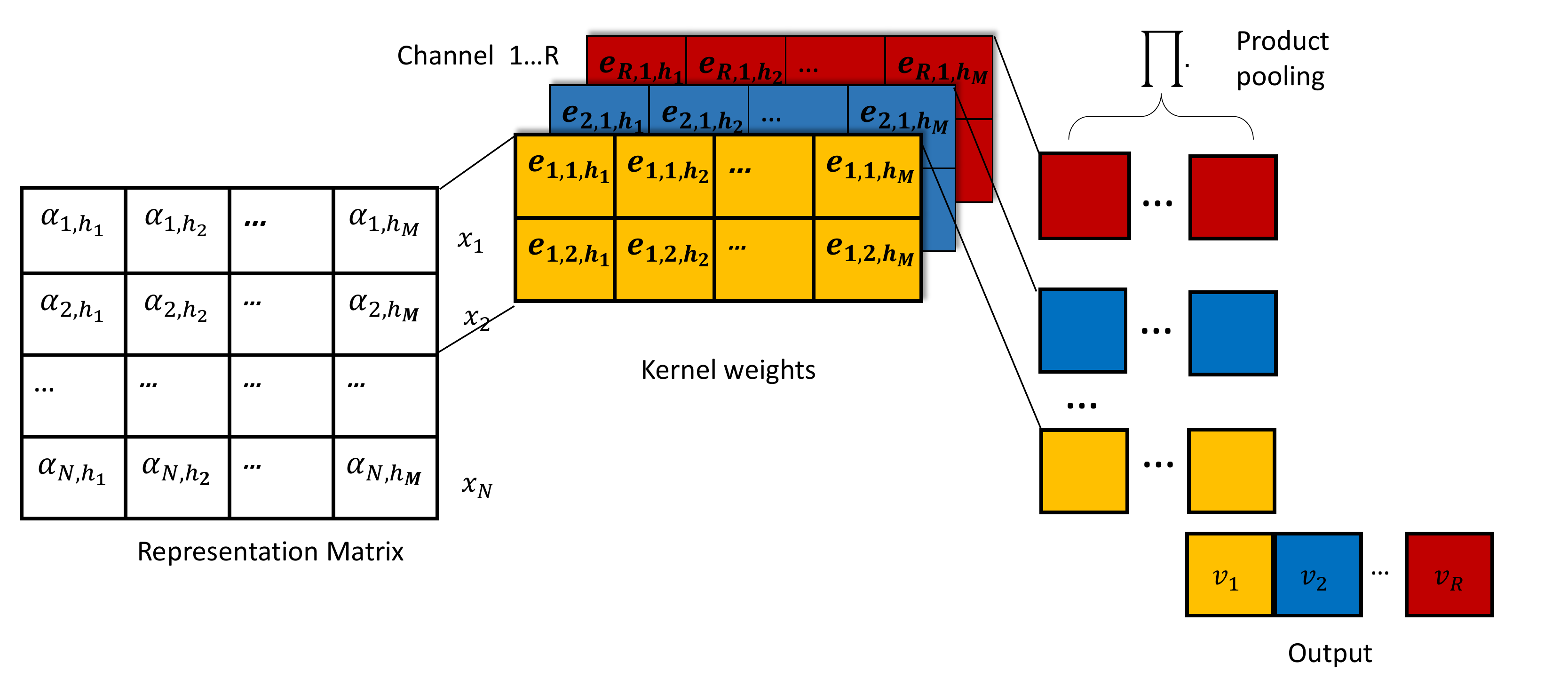}
\caption{Realization of QMWF-LM via convolution neural network with product pooling}
\label{fig:figure-conv}
\end{figure*}

\subsubsection{Towards Convolutional Neural Network} \label{Section:to_conv}

We will show that the projection from the global representation $\ket{\psi_S}$ to the local one $\ket{\psi_S^{ps}}$ can be realized by a Convolutional Neural Network (CNN) with product pooling~\cite{Cohen2016On}. To see this,  we can put the CP decomposition Eq. \ref{eq:tensor-decomposition} of $\mathcal{T}$ in Eq. \ref{eq:projection}, and obtain:
\begin{equation}
\label{eq:as-conv}
\braket{\psi_S^{ps}|\psi_S} =\sum_{r=1}^{R}t_{r}\prod_{i=1}^{N}(\sum_{h_i=1}^{M}e_{r,i,h_i}\cdot\alpha_{i,h_i})
\end{equation}
The above equation provides a connection between the quantum-inspired LM and the CNN design. The CNN interpretations of Eq.~\ref{eq:as-conv} are summarized in Table~\ref{tab:layers} and also illustrated in Fig. \ref{fig:figure-conv}.

Given a sequence with $N$ words, each is represented by an vector $\bm{x}_i = (\alpha_{i,h_1},\ldots,\alpha_{i,h_M})^T$. 
The convolution function is $\sum_{h_i=1}^{M}e_{r,i,h_i}\cdot\alpha_{i,h_i}$, which is the inner product $<\!\bm{x}_i, \bm{e}_{r,i}\!>$ between $\bm{x}_i$ and $\bm{e}_{r,i}$. The input vector $\bm{x}_i$ is a kind of local information and its entries $\alpha_{i,h_i}$ actually are the values in the local tensor $\mathcal{A}$. The entries in the vector $\bm{e}_{r,i}$ decomposed from the global tensor $\mathcal{T}$, are now parameters to be trained in the convolutional layer. 
Such an inner product, can be considered as a mapping from the global information $\bm{e}_{r,i}$ to the local representation $\bm{x}_i$. After that, the product pooling layer (i.e., $\prod_r$, see Table~\ref{tab:layers}) multiplies all the mapping results $\Sigma_{r,i}=<\!\bm{x}_i, \bm{e}_{r,i}\!>$ for all the $N$ words.

%This procedure can be seen as a convolutional network architecture which consists of an input layer, a convolution layer, a pooling layer and an output layer, which is shown in Table \ref{tab:layers}. 

%The input layer maps the words $x_i$ to a vector $[\alpha_{h_1},\ldots,\alpha_{h_M}]$. As shown in Eq. \ref{eq:projection}, $\alpha_i$ is the local semantic information. We will obtain a representation of our input shown in Fig. \ref{fig:figure-conv}. 

As mentioned above, a set of $N$ decomposed vectors corresponds to a subspace of the high-dimensional tensor. The rank $R$ is the number of decomposed subspace, and this number corresponds to the number of convolution channels.  In Fig. \ref{fig:figure-conv}, different color means different channels of convolution. 
Following the input layer, a convolution layer with $R$ channels calculates weighted sums of the representation vectors $\bm{x}_i$ and the vectors $\bm{e}_{r,i}$ (as kernel weights). In Eq.~\ref{eq:as-conv}, one can sum $R$ products $\Pi_{r}$ with weights $t_r$ to obtain the projection $\braket{\psi_S^{ps}|\psi_S}$. Then, a vector $\bm{v}=(v_1,\ldots,v_R)^T$ can be used to represent a sequence of words, where $v_r={t_r}\cdot\Pi_{r}$ $(r=1,\ldots,R$).

\begin{comment}
It's worth noting that the kernel weights is unshared in general. The decomposition of a \textit{symmetric} tensor can make the unit vectors $\bm{e}_r$ as same as each order, which means $\bm{e}_r=\bm{e}_{r,1}=\bm{e}_{r,2}=\ldots=\bm{e}_{r,N}$. This is shared case that is similar to the traditional CNN model. As shown in Eq. \ref{eq:tensor-decomposition}, $R$ is the rank of $\mathcal{T}$ which is defined as the smallest number of rank-one tensors that generate $\mathcal{T}$ as their sum. In Eq. \ref{eq:as-conv}, $R$ is corresponding with the channels of convolution and $\bm{e}_{r,i}$ is the weights of global semantic information. We can see that in QMWF-LM, convolution neural network can extract the principal semantic information in tensor space.
\end{comment}

\begin{table}[t]
\caption{CNN Interpretation of Projection}
\label{tab:layers}
\centering
\begin{tabular}{ | l | c |}
  \hline
  Input & $ \bm{x}_i = (\alpha_{i,h_1},\ldots,\alpha_{i,h_M})^T$\\ \hline
  Convolution  & $\Sigma_{r,i} = \sum_{h_i=1}^{M}e_{r,i,h_i}\cdot\alpha_{i,h_i}$ \\ \hline
  Product pooling & $\Pi_{r} = \prod_{i=1}^{N}\Sigma_{r,i}$\\ \hline  
  Output & $\sum_{r=1}^{R}t_{r}\cdot\Pi_{r}$ \\
  \hline
\end{tabular}
\end{table}

%Recall to our prejection process, if we let $v_r={t_r}\cdot\Pi_{r}$ represents  the principal information of high-dimensional tensor. Then we can use $\bm{v}=(v_1,\ldots,v_R)^T$ for a representation of a sentence.

% It's worth noting that the decomposition of a \textit{symmetric} tensor can make the unit vectors $\bm{e}_r$ as same as each order, which means $\bm{e}_r=\bm{e}_{r,1}=\bm{e}_{r,2}=\ldots=\bm{e}_{r,N}$. The projection in Eq.~\ref{eq:as-conv} can be simplified to:
% \begin{equation}
% \label{eq:as-conv-sharing}
% \braket{\psi_S^{ps}|\psi_S}=\sum_{r=1}^{R}t_{r}\prod_{i=1}^{N}(\sum_{h_i=1}^{M}e_{r,h_i}\cdot\alpha_{i,h_i})
% \end{equation}
% In this way, we will get a important property about convolutional neural networks, i.e., the weight sharing~\cite{Cohen2016On}.

% This process is significant. It means that in tensor decomposition, \zhan{convolution can approximate the CP decomposition.} For quantum many-body language model, the semantic space constructed by tensor is large. Convolution will extracts the most import semantic informations from this space.

\subsubsection{Algorithm} \label{section:algorithm}
Based on the above ideas, a practical algorithm can be obtained with four parts as follows, also shown in Fig. \ref{fig:figure-conv}:

\begin{itemize}
\item \textbf{Input layer}

The input to our model, a sequence of words $S$, is composed of $N$ words or patches $[x_1,...,x_N]$. Each word $x_{i}$ will be represented by a vector $(\alpha_{i,h_1},...,\alpha_{i,h_M})^T$. Then, we will get a representation matrix ${\bf S} \in\mathbb{R}^{N\times M}$.
% \zhan{revise}

\item \textbf{Convolution layer}
For each word or patch, the convolution is computed as follows: $\Sigma_{r,i} = \sum_{h_i = 1}^Me_{r,i,h_i}\cdot\alpha_{i,h_i}$ ($r=1,\ldots,R$), where $R$ is the number of convolution channels.

\item \textbf{Product pooling layer}
We apply the product pooling on the results of the convolution. It multiplies a number ($N$) of $\Sigma_{r,i}$ to get the $\Pi_{r} = \prod_{i=1}^{N}\Sigma_{r,i}$, where $\Pi_{r}\in\mathbb{R}$.

\item \textbf{Output}
Finally, we represent a sequence $S$ using a vector $\bm{v} = (v_1,\ldots,v_R)^T\in\mathbb{R}^R$.\end{itemize}

%Recalling the example introduced in Section \ref{eq:singlewave_index}. $\ket{\phi_1}\otimes\ket{\phi_3}$ is the \textit{compound meaning} which means the interaction information between word $x_1$ and word $x_2$, The coefficients of basises $\alpha_{1,1} \alpha_{2,3})$ is the probability with $\ket{\phi_1}\otimes\ket{\phi_3}$. The projection from global to local representation is a probability measurement. Since we utilize the convolutional neural network to realize a projection. The interaction information between $x_1$ and $x_2$ is modeled by adopting the convolution operation on the coefficients of these basises.

The above algorithm can model the sentence representation and can be applied in natural language processing tasks such as classification or matching between sentence pairs. It is worth noting that the decomposition of a \textit{symmetric} tensor can make the unit vectors $\bm{e}_r$ as same as each order, which means for a convolutional kernel, $\bm{e}_{r,1}=\bm{e}_{r,2}=\ldots=\bm{e}_{r,N}$. 
In this way, we will get a property about convolutional neural networks, i.e., the weight sharing.

\section{Applications}
\label{sec:appQA}
%As introduced in section\ref{section:QMWF-framework}, We utilize the quantum many-body wave function to model the texts and apply this to question answering tasks.

Question Answering (QA) tasks aim to rank the answers from a candidate answer pool, given a question. The ranking is based on the matching scores between question and answer sentences.
The key points are to build effective representations of the question and answer sentences and measure their matching score over such representations.
In this paper, we model the question and answer sentences with quantum many-body wave functions and apply the algorithm in Section~\ref{section:algorithm} to obtain question and answer sentences and match pairs of them.

Compared with the ad-hoc retrieval task, the question in the QA task is usually a piece of fluent natural language instead of a phrase or multiple keywords. The candidate answers are also shorter than the documents in ad-hoc retrieval. There is often less number of overlapping words between the question and answer sentences in the QA task, where semantic matching via neural network is widely used. In this paper, we apply the proposed QMWF based LM with neural network implementation in QA tasks. It should be noted that our model is also applicable to other LM based ranking tasks.

\subsection{Quantum Many-Body Representation}
As introduced in Section \ref{section: text2QLM}, each word vector $\ket{x}$ locates at the $M$ dimensional Hilbert space. The product state representation of a specific sentence is represented by the wave function in Eq.~\ref{eq:ps_state}, and the global representation of an arbitrary sentence with the same length is using the wave function in Eq.~\ref{eq:text-rep}. As introduced in Section \ref{section:product-state}, the projection onto the product state of a sentence is formulated in Eq.\ref{eq:projection}. The process of projection can be implemented by a convolution and a product pooling which has been introduced in Section \ref{Section:to_conv}.
As a text has different granularity, we can utilize two kinds of methods to obtain our input matrix ${\bf S}$.
% \zhan{revise}

\textbf{\emph{Word-level Based Input}}. We can utilize the expressive ability of pre-trained embedding. We think each dimension of word embedding is corresponding to a latent concept and a basis vector. Let $\alpha_{i,{h_i}}$ be the coefficient with respect to the corresponding basis vector, reflecting that the words can reside in specific concept with a certain probability. Given a sentence $S = [x_1,\ldots,x_N]$, where $x_i$ is a single word represented by a vector $(\alpha_{i,1},\ldots,\alpha_{i,M})^T$. Then, this sentence can be represented as a matrix in ${\bf S} \in\mathbb{R}^{N\times M}$.

\textbf{\emph{Character-level Based Input}}. Inspired by the work~\cite{kim2016character} that using character-level convolution. A sentence is represented by a sequence of characters: $S^c = [x_1,...,x_{\overline{N}}]$. The first part is a look-up table. We utilize the CNN with max pooling to obtain the representation in word level. We define the matrix $
\mathcal{Z} = [z_1,...,z_{\overline{N}}]$ as a input matrix where each column contains a vector $z_m\in R^{dk}$ that is the concatenation of a sequence of $k$ char embeddings. $N = \overline{N} - k + 1$, $d$ is the dimension of char embedding and $k$ is the window size of convolution. Each output of convolution is $(\alpha_{i,h_1},\ldots,\alpha_{i,h_M})^T$. Then, we can obtain the input matrix  ${\bf S}\in\mathbb{R}^{N\times M}$.

Based on word-level and character-level, a representation matrix ${\bf S}$ can be obtained. Then, a sentence is represented by $\bm{v} = (v_1,\ldots,v_R)^T\in\mathbb{R}^R$ based on the algorithm in Section~\ref{section:algorithm}.

\subsection{Matching via Many-body Wave Function Representation on QA}

Let $Q = [q_1,q_2\ldots q_{N_Q}]$ and $A = [a_1,a_2 \ldots a_{N_A}]$ be the sentence of the question and answer, respectively. As introduced above, we have:
% \zhan{The $\braket{\phi^{ps}_Q|\psi_S^{QA}}$ and $\braket{A_{ps}|\psi_S^{QA}}$ is projection of $\ket{\phi^{ps}_Q}$ on PS question and answer which means the probablity.} As introduced in section \ref{Section:ten-dec}. We use tensor decomposition and convolution to approximate this processing. Then we get $\braket{Q_{ps} | \psi_S^{QA}} =\sum_{r=1}^{R}t_{r}\prod_{i=1}^{N}(\sum_{h_i=1}^{M}e_{r,h_i}\cdot\alpha_{i,h_i})$ where $e_{r,h_i}$ is the convolution weights and $\alpha_{i,h_i}$ is embedding representation of quesitons. $\sum$ is convolution process, $\prod$ is the product pooling. Then we get the probability distributed representation of question
\begin{equation}
\label{eq:qd}
v^q_{i} = t_r\cdot\prod_{i=1}^{N}(\sum_{h_i=1}^{M}e_{r,i,h_i}\cdot\alpha^q_{i,h_i})
\end{equation}

\begin{equation}
\label{eq:ad}
v^a_{i} = t_r\cdot\prod_{i=1}^{N}(\sum_{h_i=1}^{M}e_{r,i,h_i}\cdot\alpha^a_{i,h_i})
\end{equation}
Then, we have
$\bm{v}^q = (v^q_{1},\ldots,v^q_{R})^T$ and $\bm{v}^a = (v^a_{1},\dots,v^a_{R})^T$ for the vector representation of the question and answer, respectively. The matching score is defined as the \textit{projection} from the answer state to the question state, which is an inner product $\braket{\bm{v}^q,\bm{v}^a}$.

\section{Literature Review}

Quantum theory is one of the most remarkable discoveries in Science. It not only can explain the behavior and movement of microscopic particles or electrons, but also have been widely applied in the macro-world problems. For instance, the quantum theory has been applied in social science and economics~\cite{Khrennikov13}, cognition and decision making~\cite{Busemeyer12,BRUZA2015383}, language model~\cite{emnlp2017qlm}, natural language processing~\cite{blacoe2013quantum,emnlp2017qlm}, and information retrieval~\cite{van2004geometry,sordoni2013modeling,aaaiZhang18}.These research directions are making use of the mathematical formulation and non-classical probability measurement of quantum theory, not necessarily for quantum computation.

%=======
%\pz{Plan to put the related work later, a section before the experiments.}
%Quantum theory is one of the most remarkable discoveries in science. It not only can  explain the behavior and movement for microscopic particles or electrons, but also can be widely applied in the macro-world problems. For instance, the quantum theory has been applied in social science and economics~\cite{Khrennikov13}, cognition and decision making~\cite{Busemeyer12,BRUZA2015383}, natural language processing~\cite{blacoe2013quantum,emnlp2017qlm}, and information retrieval~\cite{van2004geometry,sordoni2013modeling,aaaiZhang18}. For these research directions, the methodology is to make use of the mathematical formulation and non-classical probability measurement, rather than aiming at the quantum computation purpose.
%>>>>>>> 7a46b2eb4e5aeb7f7071b9d5b19395d6513a1430

In Information Retrieval (IR), van Rijsbergen for the first time proposed to adopt the mathematical formalism of quantum theory to unify various IR formal models and provide a theoretical foundation for developing new models~\cite{van2004geometry}. Later,  a number of research efforts have been made to build quantum-like IR models~\cite{PiwowarskiFLR10,DBLP:conf/ecir/ZucconA10,zhao2011novel,sordoni2013modeling}.  The main inspiration is rooted on the quantum theory as a principled framework for manipulating vector spaces and measuring probability in \textit{Hilbert space}~\cite{Melucci2011}. Piwowarski et al.~\cite{PiwowarskiFLR10} proposed a quantum IR framework, which represents the queries and documents as density matrices and subspaces, respectively. The information need space can be constructed by the tensor product of term vectors, based on a so-called multi-part system, which corresponds to a \textit{product state} (see Section~\ref{section:QMWF-framework}) of the quantum many-body wave function. However, the issue of the probability amplitudes (forming a high dimensional tensor) have not been addressed systematically. In addition, this framework has not shown the effect of tensor decomposition and its connection with the neural network design.

%, which we will address step by step in this paper.

Some quantum-inspired retrieval models are based on the \textit{analogies} between IR problems and quantum phenomena. For instance, by considering the inter-document dependency as a kind of quantum interference phenomena, a Quantum Probability Ranking Principle was proposed~\cite{DBLP:conf/ecir/ZucconA10}. In addition,  a quantum-inspired re-ranking method was developed by considering the ranking problem as a filtering process in the photon polarization~\cite{zhao2011novel}. These models are novel in terms of their quantum-inspired intuitions. In practice, they adopted the relevance scores of classical retrieval models as the input probabilities, without actually performing a quantum probability measurement (e.g., the projection measurement). 

%Therefore, these models actually do not implement the quantum probability measurement~\cite{sordoni2013modeling}.

%Quantum language modeling (QLM) for the first time develops a quantum probability based Language Model~\cite{sordoni2013modeling}. The probability uncertainties of both single and compound words are measured by the projection measurement in Hilbert space.
%For a text (a query or a document), a density matrix is then estimated based on a  Maximal Likelihood Estimation (MLE) solution. QLM can model the word dependency through encoding the uncertainty of word combinations in the density matrix, without resorting to increase the dimensions (i.e., the size of the vocabulary).
%QLM shows an effective performance on the ad-hoc retrieval task. Furthermore, using the idea of quantum entropy minimization in QLM, \citeauthor{sordoni2014learning} proposed to learn latent concept embeddings for query expansion in a supervised way~\cite{sordoni2014learning}.
%In addition,  to capture the dynamic information need in search sessions, an adaptive QLM was built with an evolution process of the density matrix~\cite{li2015modeling}. More recently, an Neural Network based QLM (NNQLM) has been built, which extends the original QLM to an end-to-end mechanism. Then, the representation and matching processes can be carried out in an effective joint training strategy. On Question Answering tasks, NNQLM shows a significant improvement on the original QLM~\cite{aaaiZhang18}.

Recently, Sordoni et al. \cite{sordoni2013modeling} proposed a principled Quantum Language Model (QLM), which generalizes the traditional statistical LM with the quantum probability theory. In QLM, the probability uncertainties of both single and compound words are measured by the \textit{projection measurement} in Hilbert space.
For a text (a query or a document), a density matrix is then estimated based on a Maximal Likelihood Estimation (MLE) solution. Practically,  QLM shows an effective performance on the ad-hoc retrieval task. Extending QLM with the idea of quantum entropy minimization, \citeauthor{sordoni2014learning} proposed to learn latent concept embeddings for query expansion in a supervised way~\cite{sordoni2014learning}.
In addition,  to capture the dynamic information need in search sessions, an adaptive QLM was built with an evolution process of the density matrix~\cite{li2015modeling}. More recently, an End-to-End Quantum-like Language Model (named as NNQLM)~\cite{aaaiZhang18} has been proposed, which built a quantum-like LM into a neural network architecture and showed a good performance on QA tasks.

In this paper, we aim to tackle these challenging problems (as identified in the Introduction) of the existing quantum-inspired LMs. Our work is inspired by the recent multidisciplinary research findings across  \textit{quantum mechanics} and \textit{machine learning} ~\cite{biamonte2016quantum} (especially neural network~\cite{Science2017,LevineYCS17}). The two different disciplines, while seemingly to have huge gaps  at the first glance, can benefit each other based on rigorous mathematical analysis and proofs. For instance, the neural network can help yield a more efficient solution for the quantum many-body problem~\cite{Science2017}. The quantum many-body system, on the other hand, can help better explain the mechanism behind the neural network~\cite{biamonte2016quantum,LevineYCS17}. The neural network based approaches have been shown effective in both the neural IR~\cite{DBLP:conf/cikm/GuoFAC16,DBLP:conf/sigir/DehghaniZSKC17,DBLP:conf/sigir/CraswellCRGM17} and QA fields~\cite{hu2014convolutional,kim2014convolutional,yin2015abcnn}. In this paper, we propose a novel quantum-inspired language modeling approach and apply it in ranking-based QA tasks. We expect that our attempt would potentially open a door for the consequent research across the fields of information retrieval, neural network, and quantum mechanics.

% and even help establish a more explainable network design~\cite{ricks2004training,stoudenmire2016supervised,LevineYCS17}.

\begin{comment}
Specifically, in this paper, we aim to develop a quantum many-body wave function inspired language modeling approach, which plays a key role in tackling the aforesaid challenges. As the neural network based approaches have been shown effective in both the neural IR~\cite{DBLP:conf/cikm/GuoFAC16,DBLP:conf/sigir/DehghaniZSKC17,DBLP:conf/sigir/CraswellCRGM17,DBLP:conf/sigir/RenCRWMR17} and QA fields~\cite{hu2014convolutional,kim2014convolutional,yin2015abcnn,qiu2015convolutional}, we expect that our attempt would potentially open a door for the consequent research across the fields of information retrieval, neural network, and quantum mechanics.
\end{comment}

%\subsection{Reconsideration of Convolution Network in Quantum Many-body function}

%We can see that the final model is similar to the convolution neural network language model except the pooling strategy. But our motivation is based on the approximation of many-body wave function. As introduced in Section , convolution is just a kind of way to decompose a high-dimensional tensor. Our model is a more general framework to the convolution neural network. Additionally, the number of convolution filters is just a hyper-parameter in convolution neural network. Actually, in tensor decomposition, the rank R which is represented by num-filters represents the main R subspaces in the tensor.

\section{Experiments}

\subsection{Datasets}
% introduce the TRECQA and WIKIQA datasets
We conduct our evaluation on three widely used datasets (summarized in Table~\ref{tab:Statistics}) for the question answering task.

\begin{itemize}
\item \textbf{TRECQA} is a benchmark dataset used in the Text Retrieval Conference (TREC)'s QA track(8-13)~\cite{wang2007jeopardy}. It contains two training sets, namely TRAIN and TRAIN-ALL. We use TRAIN-ALL, which is a larger and contains more noise, in our experiment, in order to verify the robustness of the proposed model.

\item \textbf{WIKIQA} is an open domain question-answering dataset released by Microsoft Research~\cite{yang2015wikiqa}. We remove all questions with no correct candidate answers.

\item \textbf{YahooQA}, collected from Yahoo Answers, is a benchmark dataset for community-based question answering. It contains 142627 questions and answers. The answers are generally longer than those in TRECQA and WIKIQA. As introduced in \cite{tay2017learning}, we select the QA pairs containing questions and the best answers of length 5-50 tokens after removing non-alphanumeric characters. For each question, we construct negative examples by randomly sampling 4 answers from the set of answer sentences.
\end{itemize}
\begin{table*}[pthb]\small
  \caption{Statistics of Datasets }
  \label{tab:Statistics}
  \centering
  \begin{tabular}{llllllllll}
    \hline
    \multirow{2}{*}{}& \multicolumn{3}{c}{TREC-QA} & \multicolumn{3}{c}{WIKIQA} & \multicolumn{3}{c}{YahooQA}\\
    \cline{2-10}
    &TRAIN&DEV&TEST&TRAIN&DEV&TEST&TRAIN&DEV&TEST\\
    \hline
    \#Question              & 1229& 82  & 100                  & 873     & 126& 243&56432&7082&7092\\
    \#Pairs & 53417& 1148 & 1517& 8672&1130&2351&287015&35880&35880\\
    \%Correct  & 12.0 & 19.3      & 18.7 & 12.0&12.4&12.5&20&20&20\\

    \hline
  \end{tabular}
\end{table*}

\subsection{Algorithms for Comparison}

QMWF-LM is a quantum inspired language model. The closest approaches to our QMWF-LM are QLM~\cite{sordoni2013modeling} and NNQLM~\cite{aaaiZhang18}. We treat them as our baselines. 
% It is noting that the aim of our approach is not to improve the CNN model. We focus on the QLM, So we don't compare our model with some other deeplearning models.

\begin{itemize}
\item \textbf{QLM}. The question and answer sentences are represented by the density matrices $\bm{\rho}_q$ and $\bm{\rho}_a$, respectively. Then the score function is based on the negative Von-Neumann (VN) Divergence between $\bm{\rho}_q$ and $\bm{\rho_a}$.
\item \textbf{NNQLM-\uppercase\expandafter{\romannumeral2}}. This model is an end-to-end quantum language model. We actually compare our model with NNQLM-\uppercase\expandafter{\romannumeral2}, which performs much better than NNQLM-\uppercase\expandafter{\romannumeral1}~\cite{aaaiZhang18}. The question and answer sentences are also encoded in the density matrix, but with the embedding vector as the input. The density matrix $\bm{\rho}$ is trained by a neural network. The matching score is computed by the convolutional neural network over the joint representation of two density matrices $\bm{\rho}_q$ and $\bm{\rho}_a$.
\item \textbf{QMWF-LM}. QMWF-LM is the model introduced in this paper. It is inspired by the quantum many-body wave function. \textit{QMWF-LM-word} is the model whose input matrix is the word embedding. \textit{QMWF-LM-char} is the model whose input matrix is based on char embedding.
\end{itemize}

Since we utilize the CNN with product pooling to implement the QMWF based LM, we compare our model with a range of CNN-based QA models~\cite{severyn2015learning,severyn2016modeling,dos2016attentive}. Additional CNN-based models include QA-CNN \cite{dos2016attentive}, and AP-CNN which is the attentive pooling network. Our focus is to show the potential of the QMWF-inspired LM and its connection with CNN, rather than a comprehensive comparison with all the recent CNN based QA models. Therefore, we just pick up a couple of basic and typical CNN-based QA models for comparison.

%Some other deep learning model based on recurrent neural network (RNN) is not be compared in our experiment.

\subsection{Evaluation Metrics}

For experiments on TRECQA and WIKIQA, we use the same matrix as in \cite{severyn2015learning}, namely the  MAP~(mean average precision) and MRR~(mean reciprocal rank). For experiments on YahooQA dataset, we use the same metrics as in \cite{wan2016deep}, namely Precision@1~(P@1) and MRR. P@1 is defined by $\frac{1}{N} \sum_{1}^N[rank(A^*)=1]$ where [$\cdot$] is the indicator function and $A^*$ is the ground truth.

According to Wilcoxon signed-rank test, the symbols $\alpha$ and $\beta$ denote the statistical significance (with $p<0.05$) over QLM and NNQLM-\uppercase\expandafter{\romannumeral2}, respectively, in experimental table.
% where the symbols $\alpha$ and $\beta$ denote the statistical significance (according to Wilcoxon signed-rank test) over QLM and 

\subsection{Implementation Details and Hyperparameters}

For QLM and NNQLM, we use the same parameters introduced in \cite{aaaiZhang18}.
The model is implemented by Tensorflow and the experimental machine is based on TITAN X GPU. We train our model for 50 epochs and use the best model obtained in the dev dataset for evaluation in the test set.  We utilize the Adam \cite{kingma2014adam} optimizer with learning rate [0.001,0.0001,0.00001]. The batch size is tuned among [80,100,120,140]. The L2 regularization is tuned among [0.0001,0.00001,0.000001]. For QMWF-LM-word, we initialize the input layer with 300-dimensional Glove vectors \cite{pennington2014glove}. For QMWF-LM-char, the initial char embedding is a one-hot vector. In QMWF algorithm, we use the logarithm value for the product pooling, and use two or three words in a patch to capture the phrase information.

%In order to make our algorithm work and improve the experimental results, we did some changes in our train process. In our paper, the parameters of our model is between 0 and 1, and product pooling strategy may cause the value field to change because of the multiplication operation. So in our experiment, we take the logarithm of the product state. Then as for the input layer introduced in \ref{section:algorithm}, a single word $x_i$ is a patch in our convolution feature map. In train process, we extends the patch to two or three words to capture the phrase information in sentence pairs and this is similar to the CNN deep learning model.

% $
% c_{pooled}=[pool(\Sigma_{r,i}),...,pool(\Sigma_{r,N}) ], 
% $ where $pool(\Sigma_{r,i})$
  
\subsection{Experimental Results}

% \begin{table}[thbp]\footnotesize
%   \caption{Results on TREC-QA and WIKIQA}
%   \label{tab:Deep Learning methods}
%   \centering
%   \begin{tabular}{lllll}
%     \hline
%     \multirow{2}{*}{Method}& \multicolumn{2}{c}{TREC-QA} & \multicolumn{2}{c}{WIKIQA}\\
%     \cline{2-5}
%     &MAP&MRR&MAP&MRR\\
%      \hline
%     % WordCnt              & 0.6303& 0.6562                         & 0.4837& 0.4852 \\
%     % WgtWordCnt & 0.6626& 0.6991& 0.5024& 0.5044\\
%     % \hline
%     QLM    & 0.6784            & 0.7265 &0.5109  &0.5148 \\
%     QLM$_{T}$    & 0.6683 & 0.7280 &0.5108  &0.5145\\
%     \hline
%     NNQLM-\uppercase\expandafter{\romannumeral1}  & 0.6791 & 0.7529    &0.5462  &0.5574 \\
%     NNQLM-\uppercase\expandafter{\romannumeral2}  & \textbf{0.7589} & \textbf{0.8254}   &0.6496  &0.6594 \\
%     QA-CNN & -&-&-& -\\
%      \hline
%      QMWF-LM & 0.756 & 0.814 & 0.695 & 0.710 \\
%     \hline
% \end{tabular}
% \end{table}

\begin{table}[t]

\caption{Experimental Result on TRECQA (raw). $\alpha$ denotes significant improvement over QLM.}
\label{tab:trecqa}
\centering
\begin{tabular*}{8cm}{lll}
\hline

Model & MAP & MRR \\
\hline
QLM & 0.678 & 0.726\\
NNQLM-\uppercase\expandafter{\romannumeral2}  & 0.759 & 0.825\\
\hline

CNN (Yu et al.) \cite{yu2014deep} & 0.711 & 0.785\\
CNN (Severyn) \cite{severyn2016modeling} &0.746 & 0.808\\
aNMM (Yang et al.) \cite{yin2015abcnn} & 0.749 & 0.811\\
% QA-CNN (max) & 0.734 $^{\alpha}$ & 0.790 $^{\alpha}$\\
% QA-CNN (mean) & 0.756 $^{\alpha}$ & 0.814 $^{\alpha}$\\
\hline
QMWF-LM-char & 0.715$^{\alpha}$ & 0.758 $^{\alpha}$\\
QMWF-LM-word & \textbf{0.752} $^{\alpha}$ & \textbf{0.814} $^{\alpha}$ \\

\hline
\end{tabular*}
\end{table}

Table \ref{tab:trecqa} reports the results on the TRECQA dataset. The first group shows a comparison of three quantum inspired language models.  QMWF-LM-word significantly outperforms QLM by 10.91\% on MAP and 12.12\% on MRR, respectively. The result  of QMWF-LM-word is comparable with that of NNQLM-\uppercase\expandafter{\romannumeral2}. In the second group, we compare our model with a range of CNN-based models against their results reported in the corresponding original papers. We can see that the QMWF-LM-word improves the CNN model in~\cite{yu2014deep} by 5.77\% on MAP, and 3.69\% on MRR, respectively. % Moreover, QMWF-LM outperforms QA-CNN (max) by 2\% on MAP and 2\% on MRR. The result is also comparable to QA-CNN (mean).

%We also evaluate the QA-CNN model with mean pooling which represented by QACNN* in our table.
Table \ref{tab:wikiqa} reports the results on WIKIQA. QMWF-LM-word significantly outperforms QLM by 35.74\% on MAP, and 37.86\% on MRR, as well as  NNQLM-\uppercase\expandafter{\romannumeral2} by 6.92\% on MAP, and 7.74\% on MRR. In comparison with CNN models, QMWF-LM-word outperforms QA-CNN and AP-CNN by (1\%$\sim$2\%) on both MAP and MRR, based on their reported results. 
% It outperforms the QA-CNN (max) by 4\% on MAP and 5\% on MRR and achieves a competitive result compared with QA-CNN (mean).

The experimental results on YahooQA are shown in Table \ref{tab:yahooqa result}. Our QMWF-LM-word achieves a significant improvement over QLM by 45.57\% on P@1 and 23.34\% on MRR, respectively. It also outperforms NNQLM-\uppercase\expandafter{\romannumeral2}  on P@1  by 23.39\% and on MRR by 10.70\%, respectively. Compared with the results of other CNN models on YahooQA dataset as reported  in~\cite{tay2017learning}, QMWF-LM-word shows improvements over AP-CNN by about 2.67\% on P@1 and 2.61\% on MRR, respectively. Note that the data preprocessing of YahooQA dataset in our experiments is a little different, as we randomly sample four negative examples from the answers sentence set.

As we can see from the results, the performance of QMWF-LM-char is not as good as that of QMWF-LM-word. However, compared with NNQLM-\uppercase\expandafter{\romannumeral2}, QMWF-LM-char has better results on WIKIQA and YahooQA datasets. We will give a further analysis of this phenomenon in Section~\ref{sec:compare_w_c}. 
% The QA-CNN (max) outperforms the QMWF-LM by 2\% on  P@1 and 1\% on MRR. Additionally, we can see that QMWF-LM is comparable with QA-CNN (mean). 

% ########################## analusis my rank

\begin{table}[t]
\caption{Experimental Result on WIKIQA. $\alpha$ and $\beta$ denote significant improvement  over QLM and NNQLM-\uppercase\expandafter{\romannumeral2}, respectively.}
\label{tab:wikiqa}
\centering
\begin{tabular*}{8cm}{lll}
\hline

Model & MAP & MRR \\
\hline
QLM & 0.512 & 0.515\\
NNQLM-\uppercase\expandafter{\romannumeral2}  &0.650  &0.659 \\
\hline
QA-CNN (Santos et al.) \cite{dos2016attentive} & 0.670 & 0.682\\
AP-CNN (Santos et al.) \cite{dos2016attentive}& 0.688 & 0.696\\
% QA-CNN (max) & 0.651 ${^\alpha}$ & 0.660 ${^\alpha}$ \\
% QA-CNN (mean) & 0.684 $^{\alpha\beta}$ & 0.698 $^{\alpha\beta}$\\

\hline
QMWF-LM-char & 0.657 $^{\alpha}$ & 0.679 $^{\alpha}$ \\
QMWF-LM-word & \textbf{0.695} $^{\alpha\beta}$ & \textbf{0.710} $^{\alpha\beta}$  \\
\hline
\end{tabular*}
\end{table}

\begin{table}[t]
\caption{Experimental Result on YahooQA. $\alpha$ and $\beta$ denote significant improvement  over QLM and NNQLM-\uppercase\expandafter{\romannumeral2}, respectively.}
\label{tab:yahooqa result}
\centering
\begin{tabular*}{8cm}{lll}
\hline

Model & P@1 & MRR \\
\hline
Random guess & 0.200 & 0.457 \\
QLM & 0.395 & 0.604\\
NNQLM-\uppercase\expandafter{\romannumeral2}  &0.466  &0.673 \\
\hline
QA-CNN (Santos et al.)\cite{dos2016attentive}&0.564&0.727\\
AP-CNN (Santos et al.)\cite{dos2016attentive} & 0.560&0.726\\
% QA-CNN (max) & 0.591 $^{\alpha\beta}$& 0.752 $^{\alpha\beta}$\\
% QA-CNN (mean) & 0.578 $^{\alpha\beta}$ & 0.746 $^{\alpha\beta}$\\
\hline
QMWF-LM-char & 0.513 $^{\alpha\beta}$ & 0.696 $^{\alpha\beta}$ \\
QMWF-LM-word &\textbf{0.575} $^{\alpha\beta}$ & \textbf{0.745} $^{\alpha\beta}$ \\

\hline
\end{tabular*}
\end{table}
\subsection{Discussion and Analysis}
\label{sec:analysis}

\subsubsection{The Result Analysis}

The experimental results show that our proposed model, namely QMWF-LM, has achieved a significant improvement over QLM on three QA datasets, and outperforms NNQLM-\uppercase\expandafter{\romannumeral2} on both WIKIQA and YahooQA datasets. Especially on YahooQA, which is the largest among the three datasets, QMWF-LM significantly outperforms the other two quantum-inspired LM approaches. 
Note that the original QLM is trained in an unsupervised manner. Therefore, unsurprisingly it under-performs the other two supervised models (i.e., NNQLM and  QMWF-LM). NNQLM adopts the embedding vector as its input and uses the convolutional neural network to train the density matrix. However, the interaction among words is not taken into account in NNQLM. The experiment also shows that the proposed model can achieve a comparable and even better performance over a couple of CNN-based QA models. In summary, our proposed model reveals the analogy between the quantum many-body system and the language modeling, and further effectively bridge the quantum many-body wave function inspired LM with the neural network design.

\subsubsection{The Comparison between Word embedding and Char embedding}
\label{sec:compare_w_c}
In our experiment, the input layer is based on word embedding and char embedding. For char embedding, we treat the text as a kind of raw signal which has been proved effective in modeling sentence~\cite{kim2016character}. As char embedding is initialized by one-hot vector, the semantic information is only based on training dataset. In QA tasks, the semantic matching requires a relatively large data for training the embeddings. Therefore, compared with char embedding, pre-trained word embedding trained by an external large corpus (rather than training data only) is more effective.
% In this section, we analyze some factors of our model. 
% \emph{The dimension of M} which represents the semantic space of word resided in. 

\subsubsection{Influence of Channels in Convolution}
% \begin{figure}[pthb]
% \centering
% \includegraphics[scale=0.50]{rank_R.pdf}
% \caption{The Influence of Rank}
% \label{fig:rank}
% \end{figure}

As introduced in Section~\ref{section:QMWF-framework}, the number of convolution channels is corresponding to $R$ which is the rank of a tensor. In our problem, it is not straightforward to determine the rank of a concerned tensor. We select the optimal $R$
in a range $[20, 200]$ with increment 5. For different datasets, we set a suitable number of channels to obtain the best performance. The number of channels is set to 150 for TRECQA and WIKIQA dataset, and 200 For YahooQA dataset. 

%As show in Fig.~\ref{fig:rank}, we show the influence of channels on WIKIQA dataset.

\subsubsection{Efficiency Analysis}

As we utilize convolution neural network to implement the operation of tensor decomposition. The efficiency relies on the convolution neural network. In our experiment, for QMWF-LM-char, the training epoch is set to be 200, while for QMWF-LM-word, after training 20 epochs, we will obtain the results.

\section{Conclusions and Future Work}

In this paper, we propose a Quantum Many-body Wave Function (QMWF) inspired Language Modeling (QMWF-LM) framework. We have shown that the incorporation of QMWF has enhanced the representation space of quantum-inspired LM approaches, in the sense that QMWF-LM can model the complex interactions among words with multiple meanings. In addition, inspired by the recent progress on solving the quantum many-body problem and its connection to the neural network, we bridge the gap  between the quantum-inspired language modeling and the convolutional neural network. Specifically, a series of derivations (based on projection and tensor decomposition) show that the quantum many-body language modeling representation and matching process can be implemented by the convolutional neural network (CNN) with product pooling. This result simplifies the estimation of the probability amplitudes in QMWF-LM. Based on this idea, we provide a simple algorithm in a basic CNN architecture.

We implement our approach on the question answering task. Experiments on three QA datasets have demonstrated the effectiveness of our proposed QMWF based LM. It achieves a significant improvement over its quantum-like counterparts, i.e., QLM and NNQLM. It can also achieve effective performance compared with several  convolutional neural network based approaches.
Furthermore, based on the analytical and empirical evidence presented in this paper, we can conclude that the proposed approach has made the first step to bridge the quantum-inspired formalism, language modeling and neural network structure in a principled manner.

In the future, the quantum many-body inspired language model should be investigated in more depth from both theoretical and empirical perspectives. Theoretically, a more unified framework to explain another widely-adopted neural network architecture, i.e., Recurrent Neural Network (RNN), can be explored based on the mechanism of quantum  many-body language modeling. Practically, we will apply and evaluate QMWF-LM on other IR or NLP tasks with larger scale datasets.

% Essentially, the latter (based on subspaces) can be considered as a generalization of the former (based on subsets)~\cite{Melucci2011,sordoni2013looking}.
%

%Among those approaches, statistical language modelingis widely used approach to modeling such uncertainties by computing the probabilities of single words or compound words~\cite{DBLP:series/synthesis/2008Zhai}.

\section{Acknowledgments}
This work is supported in part by the state key development program of China  (grant No. 2017YFE0111900), Natural Science Foundation of China (grant No. U1636203, 61772363), and the European Union's Horizon 2020 research and innovation programme under the Marie Sk\l{}odowska-Curie grant agreement No. 721321.

\bibliographystyle{ACM-Reference-Format}
\bibliography{sample-bibliography}

%%% -*-BibTeX-*-
%%% Do NOT edit. File created by BibTeX with style
%%% ACM-Reference-Format-Journals [18-Jan-2012].

\begin{thebibliography}{42}

%%% ====================================================================
%%% NOTE TO THE USER: you can override these defaults by providing
%%% customized versions of any of these macros before the \bibliography
%%% command.  Each of them MUST provide its own final punctuation,
%%% except for \shownote{}, \showDOI{}, and \showURL{}.  The latter two
%%% do not use final punctuation, in order to avoid confusing it with
%%% the Web address.
%%%
%%% To suppress output of a particular field, define its macro to expand
%%% to an empty string, or better, \unskip, like this:
%%%
%%% \newcommand{\showDOI}[1]{\unskip}   % LaTeX syntax
%%%
%%% \def \showDOI #1{\unskip}           % plain TeX syntax
%%%
%%% ====================================================================

\ifx \showCODEN    \undefined \def \showCODEN     #1{\unskip}     \fi
\ifx \showDOI      \undefined \def \showDOI       #1{#1}\fi
\ifx \showISBNx    \undefined \def \showISBNx     #1{\unskip}     \fi
\ifx \showISBNxiii \undefined \def \showISBNxiii  #1{\unskip}     \fi
\ifx \showISSN     \undefined \def \showISSN      #1{\unskip}     \fi
\ifx \showLCCN     \undefined \def \showLCCN      #1{\unskip}     \fi
\ifx \shownote     \undefined \def \shownote      #1{#1}          \fi
\ifx \showarticletitle \undefined \def \showarticletitle #1{#1}   \fi
\ifx \showURL      \undefined \def \showURL       {\relax}        \fi
% The following commands are used for tagged output and should be
% invisible to TeX
\providecommand\bibfield[2]{#2}
\providecommand\bibinfo[2]{#2}
\providecommand\natexlab[1]{#1}
\providecommand\showeprint[2][]{arXiv:#2}

\bibitem[\protect\citeauthoryear{Basile and Tamburini}{Basile and
  Tamburini}{2017}]%
        {emnlp2017qlm}
\bibfield{author}{\bibinfo{person}{Ivano Basile} {and} \bibinfo{person}{Fabio
  Tamburini}.} \bibinfo{year}{2017}\natexlab{}.
\newblock \showarticletitle{Towards Quantum Language Models}. In
  \bibinfo{booktitle}{\emph{Proc. of EMNLP}}. \bibinfo{pages}{1840 --1849}.
\newblock


\bibitem[\protect\citeauthoryear{Biamonte, Wittek, Pancotti, Rebentrost, Wiebe,
  and Lloyd}{Biamonte et~al\mbox{.}}{2016}]%
        {biamonte2016quantum}
\bibfield{author}{\bibinfo{person}{Jacob Biamonte}, \bibinfo{person}{Peter
  Wittek}, \bibinfo{person}{Nicola Pancotti}, \bibinfo{person}{Patrick
  Rebentrost}, \bibinfo{person}{Nathan Wiebe}, {and} \bibinfo{person}{Seth
  Lloyd}.} \bibinfo{year}{2016}\natexlab{}.
\newblock \showarticletitle{Quantum machine learning}.
\newblock \bibinfo{journal}{\emph{arXiv preprint arXiv:1611.09347}}
  (\bibinfo{year}{2016}).
\newblock


\bibitem[\protect\citeauthoryear{Blacoe, Kashefi, and Lapata}{Blacoe
  et~al\mbox{.}}{2013}]%
        {blacoe2013quantum}
\bibfield{author}{\bibinfo{person}{William Blacoe}, \bibinfo{person}{Elham
  Kashefi}, {and} \bibinfo{person}{Mirella Lapata}.}
  \bibinfo{year}{2013}\natexlab{}.
\newblock \showarticletitle{A Quantum-Theoretic Approach to Distributional
  Semantics.}. In \bibinfo{booktitle}{\emph{Proc. of HLT-NAACL}}.
  \bibinfo{pages}{847--857}.
\newblock


\bibitem[\protect\citeauthoryear{Bruza, Wang, and Busemeyer}{Bruza
  et~al\mbox{.}}{2015}]%
        {BRUZA2015383}
\bibfield{author}{\bibinfo{person}{Peter~D. Bruza}, \bibinfo{person}{Zheng
  Wang}, {and} \bibinfo{person}{Jerome~R. Busemeyer}.}
  \bibinfo{year}{2015}\natexlab{}.
\newblock \showarticletitle{Quantum cognition: a new theoretical approach to
  psychology}.
\newblock \bibinfo{journal}{\emph{Trends in Cognitive Sciences}}
  \bibinfo{volume}{19}, \bibinfo{number}{7} (\bibinfo{year}{2015}),
  \bibinfo{pages}{383 -- 393}.
\newblock


\bibitem[\protect\citeauthoryear{Busemeyer and Bruza}{Busemeyer and
  Bruza}{2013}]%
        {Busemeyer12}
\bibfield{author}{\bibinfo{person}{Jerome~R. Busemeyer} {and}
  \bibinfo{person}{Peter~D. Bruza}.} \bibinfo{year}{2013}\natexlab{}.
\newblock \bibinfo{booktitle}{\emph{{Quantum Models of Cognition and
  Decision}}}.
\newblock \bibinfo{publisher}{Cambridge University Press}.
\newblock


\bibitem[\protect\citeauthoryear{{Carleo} and {Troyer}}{{Carleo} and
  {Troyer}}{2017}]%
        {Science2017}
\bibfield{author}{\bibinfo{person}{G. {Carleo}} {and} \bibinfo{person}{M.
  {Troyer}}.} \bibinfo{year}{2017}\natexlab{}.
\newblock \showarticletitle{{Solving the quantum many-body problem with
  artificial neural networks}}.
\newblock \bibinfo{journal}{\emph{Science}}  \bibinfo{volume}{355}
  (\bibinfo{year}{2017}), \bibinfo{pages}{602--606}.
\newblock


\bibitem[\protect\citeauthoryear{Cohen, Sharir, and Shashua}{Cohen
  et~al\mbox{.}}{2016}]%
        {Cohen2016On}
\bibfield{author}{\bibinfo{person}{Nadav Cohen}, \bibinfo{person}{Or Sharir},
  {and} \bibinfo{person}{Amnon Shashua}.} \bibinfo{year}{2016}\natexlab{}.
\newblock \showarticletitle{On the Expressive Power of Deep Learning: A Tensor
  Analysis}.
\newblock \bibinfo{journal}{\emph{Computer Science}} (\bibinfo{year}{2016}).
\newblock


\bibitem[\protect\citeauthoryear{Craswell, Croft, de~Rijke, Guo, and
  Mitra}{Craswell et~al\mbox{.}}{2017}]%
        {DBLP:conf/sigir/CraswellCRGM17}
\bibfield{author}{\bibinfo{person}{Nick Craswell}, \bibinfo{person}{W.~Bruce
  Croft}, \bibinfo{person}{Maarten de Rijke}, \bibinfo{person}{Jiafeng Guo},
  {and} \bibinfo{person}{Bhaskar Mitra}.} \bibinfo{year}{2017}\natexlab{}.
\newblock \showarticletitle{{SIGIR} 2017 Workshop on Neural Information
  Retrieval (Neu-IR'17)}. In \bibinfo{booktitle}{\emph{Proc. of SIGIR}}.
  \bibinfo{pages}{1431--1432}.
\newblock


\bibitem[\protect\citeauthoryear{Dehghani, Zamani, Severyn, Kamps, and
  Croft}{Dehghani et~al\mbox{.}}{2017}]%
        {DBLP:conf/sigir/DehghaniZSKC17}
\bibfield{author}{\bibinfo{person}{Mostafa Dehghani}, \bibinfo{person}{Hamed
  Zamani}, \bibinfo{person}{Aliaksei Severyn}, \bibinfo{person}{Jaap Kamps},
  {and} \bibinfo{person}{W.~Bruce Croft}.} \bibinfo{year}{2017}\natexlab{}.
\newblock \showarticletitle{Neural Ranking Models with Weak Supervision}. In
  \bibinfo{booktitle}{\emph{Proc. of SIGIR}}. \bibinfo{pages}{65--74}.
\newblock


\bibitem[\protect\citeauthoryear{dos Santos, Tan, Xiang, and Zhou}{dos Santos
  et~al\mbox{.}}{2016}]%
        {dos2016attentive}
\bibfield{author}{\bibinfo{person}{C{\i}cero~Nogueira dos Santos},
  \bibinfo{person}{Ming Tan}, \bibinfo{person}{Bing Xiang}, {and}
  \bibinfo{person}{Bowen Zhou}.} \bibinfo{year}{2016}\natexlab{}.
\newblock \showarticletitle{Attentive pooling networks}.
\newblock \bibinfo{journal}{\emph{CoRR, abs/1602.03609}}
  (\bibinfo{year}{2016}).
\newblock


\bibitem[\protect\citeauthoryear{Guo, Fan, Ai, and Croft}{Guo
  et~al\mbox{.}}{2016}]%
        {DBLP:conf/cikm/GuoFAC16}
\bibfield{author}{\bibinfo{person}{Jiafeng Guo}, \bibinfo{person}{Yixing Fan},
  \bibinfo{person}{Qingyao Ai}, {and} \bibinfo{person}{W.~Bruce Croft}.}
  \bibinfo{year}{2016}\natexlab{}.
\newblock \showarticletitle{A Deep Relevance Matching Model for Ad-hoc
  Retrieval}. In \bibinfo{booktitle}{\emph{Proc. of CIKM}}.
  \bibinfo{pages}{55--64}.
\newblock


\bibitem[\protect\citeauthoryear{{Haven} and {Khrennikov}}{{Haven} and
  {Khrennikov}}{2013}]%
        {Khrennikov13}
\bibfield{author}{\bibinfo{person}{E. {Haven}} {and} \bibinfo{person}{A.
  {Khrennikov}}.} \bibinfo{year}{2013}\natexlab{}.
\newblock \bibinfo{booktitle}{\emph{{Quantum Social Science}}}.
\newblock \bibinfo{publisher}{Cambridge University Press}.
\newblock


\bibitem[\protect\citeauthoryear{Hitchcock}{Hitchcock}{1927}]%
        {hitchcock1927expression}
\bibfield{author}{\bibinfo{person}{Frank~L Hitchcock}.}
  \bibinfo{year}{1927}\natexlab{}.
\newblock \showarticletitle{The expression of a tensor or a polyadic as a sum
  of products}.
\newblock \bibinfo{journal}{\emph{Studies in Applied Mathematics}}
  \bibinfo{volume}{6}, \bibinfo{number}{1-4} (\bibinfo{year}{1927}),
  \bibinfo{pages}{164--189}.
\newblock


\bibitem[\protect\citeauthoryear{Hu, Lu, Li, and Chen}{Hu
  et~al\mbox{.}}{2014}]%
        {hu2014convolutional}
\bibfield{author}{\bibinfo{person}{Baotian Hu}, \bibinfo{person}{Zhengdong Lu},
  \bibinfo{person}{Hang Li}, {and} \bibinfo{person}{Qingcai Chen}.}
  \bibinfo{year}{2014}\natexlab{}.
\newblock \showarticletitle{Convolutional neural network architectures for
  matching natural language sentences}. In \bibinfo{booktitle}{\emph{Proc. of
  NIPS}}. \bibinfo{pages}{2042--2050}.
\newblock


\bibitem[\protect\citeauthoryear{Kim}{Kim}{2014}]%
        {kim2014convolutional}
\bibfield{author}{\bibinfo{person}{Yoon Kim}.} \bibinfo{year}{2014}\natexlab{}.
\newblock \showarticletitle{Convolutional neural networks for sentence
  classification}.
\newblock \bibinfo{journal}{\emph{arXiv preprint arXiv:1408.5882}}
  (\bibinfo{year}{2014}).
\newblock


\bibitem[\protect\citeauthoryear{Kim, Jernite, Sontag, and Rush}{Kim
  et~al\mbox{.}}{2016}]%
        {kim2016character}
\bibfield{author}{\bibinfo{person}{Yoon Kim}, \bibinfo{person}{Yacine Jernite},
  \bibinfo{person}{David Sontag}, {and} \bibinfo{person}{Alexander~M Rush}.}
  \bibinfo{year}{2016}\natexlab{}.
\newblock \showarticletitle{Character-Aware Neural Language Models.}. In
  \bibinfo{booktitle}{\emph{Proc. of AAAI}}. \bibinfo{pages}{2741--2749}.
\newblock


\bibitem[\protect\citeauthoryear{Kingma and Ba}{Kingma and Ba}{2014}]%
        {kingma2014adam}
\bibfield{author}{\bibinfo{person}{Diederik~P Kingma} {and}
  \bibinfo{person}{Jimmy Ba}.} \bibinfo{year}{2014}\natexlab{}.
\newblock \showarticletitle{Adam: A method for stochastic optimization}.
\newblock \bibinfo{journal}{\emph{arXiv preprint arXiv:1412.6980}}
  (\bibinfo{year}{2014}).
\newblock


\bibitem[\protect\citeauthoryear{Levine, Yakira, Cohen, and Shashua}{Levine
  et~al\mbox{.}}{2017}]%
        {LevineYCS17}
\bibfield{author}{\bibinfo{person}{Yoav Levine}, \bibinfo{person}{David
  Yakira}, \bibinfo{person}{Nadav Cohen}, {and} \bibinfo{person}{Amnon
  Shashua}.} \bibinfo{year}{2017}\natexlab{}.
\newblock \showarticletitle{Deep Learning and Quantum Entanglement: Fundamental
  Connections with Implications to Network Design}.
\newblock \bibinfo{journal}{\emph{CoRR}}  \bibinfo{volume}{abs/1704.01552}
  (\bibinfo{year}{2017}).
\newblock
\urldef\tempurl%
\url{http://arxiv.org/abs/1704.01552}
\showURL{%
\tempurl}


\bibitem[\protect\citeauthoryear{Li, Li, Zhang, and Song}{Li
  et~al\mbox{.}}{2015}]%
        {li2015modeling}
\bibfield{author}{\bibinfo{person}{Qiuchi Li}, \bibinfo{person}{Jingfei Li},
  \bibinfo{person}{Peng Zhang}, {and} \bibinfo{person}{Dawei Song}.}
  \bibinfo{year}{2015}\natexlab{}.
\newblock \showarticletitle{Modeling multi-query retrieval tasks using density
  matrix transformation}. In \bibinfo{booktitle}{\emph{Proc. of SIGIR}}. ACM,
  \bibinfo{pages}{871--874}.
\newblock


\bibitem[\protect\citeauthoryear{Melucci and van Rijsbergen}{Melucci and van
  Rijsbergen}{2011}]%
        {Melucci2011}
\bibfield{author}{\bibinfo{person}{Massimo Melucci} {and}
  \bibinfo{person}{Keith van Rijsbergen}.} \bibinfo{year}{2011}\natexlab{}.
\newblock \bibinfo{booktitle}{\emph{Quantum Mechanics and Information
  Retrieval}}.
\newblock \bibinfo{publisher}{Springer Berlin Heidelberg},
  \bibinfo{address}{Berlin, Heidelberg}, \bibinfo{pages}{125--155}.
\newblock


\bibitem[\protect\citeauthoryear{Metzler and Croft}{Metzler and Croft}{2005}]%
        {DBLP:conf/sigir/MetzlerC05}
\bibfield{author}{\bibinfo{person}{Donald Metzler} {and}
  \bibinfo{person}{W.~Bruce Croft}.} \bibinfo{year}{2005}\natexlab{}.
\newblock \showarticletitle{A Markov random field model for term dependencies}.
  In \bibinfo{booktitle}{\emph{Proc. of SIGIR}}. \bibinfo{pages}{472--479}.
\newblock


\bibitem[\protect\citeauthoryear{Mikolov, Sutskever, Chen, Corrado, and
  Dean}{Mikolov et~al\mbox{.}}{2013}]%
        {mikolov2013distributed}
\bibfield{author}{\bibinfo{person}{Tomas Mikolov}, \bibinfo{person}{Ilya
  Sutskever}, \bibinfo{person}{Kai Chen}, \bibinfo{person}{Greg~S Corrado},
  {and} \bibinfo{person}{Jeff Dean}.} \bibinfo{year}{2013}\natexlab{}.
\newblock \showarticletitle{Distributed representations of words and phrases
  and their compositionality}. In \bibinfo{booktitle}{\emph{Proc. of NIPS}}.
  \bibinfo{pages}{3111--3119}.
\newblock


\bibitem[\protect\citeauthoryear{Nielsen and Chuang}{Nielsen and
  Chuang}{2011}]%
        {Nielsen:2011}
\bibfield{author}{\bibinfo{person}{Michael~A. Nielsen} {and}
  \bibinfo{person}{Isaac~L. Chuang}.} \bibinfo{year}{2011}\natexlab{}.
\newblock \bibinfo{booktitle}{\emph{Quantum Computation and Quantum
  Information: 10th Anniversary Edition} (\bibinfo{edition}{10th} ed.)}.
\newblock \bibinfo{publisher}{Cambridge University Press},
  \bibinfo{address}{New York, NY, USA}.
\newblock
\showISBNx{1107002176, 9781107002173}


\bibitem[\protect\citeauthoryear{Pennington, Socher, and Manning}{Pennington
  et~al\mbox{.}}{2014}]%
        {pennington2014glove}
\bibfield{author}{\bibinfo{person}{Jeffrey Pennington},
  \bibinfo{person}{Richard Socher}, {and} \bibinfo{person}{Christopher
  Manning}.} \bibinfo{year}{2014}\natexlab{}.
\newblock \showarticletitle{Glove: Global vectors for word representation}. In
  \bibinfo{booktitle}{\emph{Proc. of EMNLP}}. \bibinfo{pages}{1532--1543}.
\newblock


\bibitem[\protect\citeauthoryear{Piwowarski, Frommholz, Lalmas, and van
  Rijsbergen}{Piwowarski et~al\mbox{.}}{2010}]%
        {PiwowarskiFLR10}
\bibfield{author}{\bibinfo{person}{Benjamin Piwowarski}, \bibinfo{person}{Ingo
  Frommholz}, \bibinfo{person}{Mounia Lalmas}, {and} \bibinfo{person}{Keith van
  Rijsbergen}.} \bibinfo{year}{2010}\natexlab{}.
\newblock \showarticletitle{What can quantum theory bring to information
  retrieval}. In \bibinfo{booktitle}{\emph{Proc. of CIKM}}.
  \bibinfo{pages}{59--68}.
\newblock


\bibitem[\protect\citeauthoryear{Severyn and Moschitti}{Severyn and
  Moschitti}{2015}]%
        {severyn2015learning}
\bibfield{author}{\bibinfo{person}{Aliaksei Severyn} {and}
  \bibinfo{person}{Alessandro Moschitti}.} \bibinfo{year}{2015}\natexlab{}.
\newblock \showarticletitle{Learning to rank short text pairs with
  convolutional deep neural networks}. In \bibinfo{booktitle}{\emph{Proc. of
  SIGIR}}. ACM, \bibinfo{pages}{373--382}.
\newblock


\bibitem[\protect\citeauthoryear{Severyn and Moschitti}{Severyn and
  Moschitti}{2016}]%
        {severyn2016modeling}
\bibfield{author}{\bibinfo{person}{Aliaksei Severyn} {and}
  \bibinfo{person}{Alessandro Moschitti}.} \bibinfo{year}{2016}\natexlab{}.
\newblock \showarticletitle{Modeling relational information in question-answer
  pairs with convolutional neural networks}.
\newblock \bibinfo{journal}{\emph{arXiv preprint arXiv:1604.01178}}
  (\bibinfo{year}{2016}).
\newblock


\bibitem[\protect\citeauthoryear{Song and Croft}{Song and Croft}{1999}]%
        {DBLP:conf/sigir/SongC99}
\bibfield{author}{\bibinfo{person}{Fei Song} {and} \bibinfo{person}{W.~Bruce
  Croft}.} \bibinfo{year}{1999}\natexlab{}.
\newblock \showarticletitle{A General Language Model for Information Retrieval
  (poster abstract)}. In \bibinfo{booktitle}{\emph{Proc. of SIGIR}}.
  \bibinfo{pages}{279--280}.
\newblock


\bibitem[\protect\citeauthoryear{Sordoni, Bengio, and Nie}{Sordoni
  et~al\mbox{.}}{2014}]%
        {sordoni2014learning}
\bibfield{author}{\bibinfo{person}{Alessandro Sordoni}, \bibinfo{person}{Yoshua
  Bengio}, {and} \bibinfo{person}{Jian-Yun Nie}.}
  \bibinfo{year}{2014}\natexlab{}.
\newblock \showarticletitle{Learning Concept Embeddings for Query Expansion by
  Quantum Entropy Minimization.}. In \bibinfo{booktitle}{\emph{Proc. of AAAI}},
  Vol.~\bibinfo{volume}{14}. \bibinfo{pages}{1586--1592}.
\newblock


\bibitem[\protect\citeauthoryear{Sordoni and Nie}{Sordoni and Nie}{2013}]%
        {sordoni2013looking}
\bibfield{author}{\bibinfo{person}{Alessandro Sordoni} {and}
  \bibinfo{person}{Jian-Yun Nie}.} \bibinfo{year}{2013}\natexlab{}.
\newblock \showarticletitle{Looking at vector space and language models for ir
  using density matrices}. In \bibinfo{booktitle}{\emph{Proc. of QI}}.
  Springer, \bibinfo{pages}{147--159}.
\newblock


\bibitem[\protect\citeauthoryear{Sordoni, Nie, and Bengio}{Sordoni
  et~al\mbox{.}}{2013}]%
        {sordoni2013modeling}
\bibfield{author}{\bibinfo{person}{Alessandro Sordoni},
  \bibinfo{person}{Jian-Yun Nie}, {and} \bibinfo{person}{Yoshua Bengio}.}
  \bibinfo{year}{2013}\natexlab{}.
\newblock \showarticletitle{Modeling term dependencies with quantum language
  models for IR}. In \bibinfo{booktitle}{\emph{Proc. of SIGIR}}. ACM,
  \bibinfo{pages}{653--662}.
\newblock


\bibitem[\protect\citeauthoryear{Tay, Phan, Tuan, and Hui}{Tay
  et~al\mbox{.}}{2017}]%
        {tay2017learning}
\bibfield{author}{\bibinfo{person}{Yi Tay}, \bibinfo{person}{Minh~C Phan},
  \bibinfo{person}{Luu~Anh Tuan}, {and} \bibinfo{person}{Siu~Cheung Hui}.}
  \bibinfo{year}{2017}\natexlab{}.
\newblock \showarticletitle{Learning to Rank Question Answer Pairs with
  Holographic Dual LSTM Architecture}. In \bibinfo{booktitle}{\emph{Proc. of
  SIGIR}}. ACM, \bibinfo{pages}{695--704}.
\newblock


\bibitem[\protect\citeauthoryear{Van~Rijsbergen}{Van~Rijsbergen}{2004}]%
        {van2004geometry}
\bibfield{author}{\bibinfo{person}{Cornelis~Joost Van~Rijsbergen}.}
  \bibinfo{year}{2004}\natexlab{}.
\newblock \bibinfo{booktitle}{\emph{The geometry of information retrieval}}.
\newblock \bibinfo{publisher}{Cambridge University Press}.
\newblock


\bibitem[\protect\citeauthoryear{Wan, Lan, Guo, Xu, Pang, and Cheng}{Wan
  et~al\mbox{.}}{2016}]%
        {wan2016deep}
\bibfield{author}{\bibinfo{person}{Shengxian Wan}, \bibinfo{person}{Yanyan
  Lan}, \bibinfo{person}{Jiafeng Guo}, \bibinfo{person}{Jun Xu},
  \bibinfo{person}{Liang Pang}, {and} \bibinfo{person}{Xueqi Cheng}.}
  \bibinfo{year}{2016}\natexlab{}.
\newblock \showarticletitle{A Deep Architecture for Semantic Matching with
  Multiple Positional Sentence Representations.}. In
  \bibinfo{booktitle}{\emph{Proc. of AAAI}}. \bibinfo{pages}{2835--2841}.
\newblock


\bibitem[\protect\citeauthoryear{Wang, Smith, and Mitamura}{Wang
  et~al\mbox{.}}{2007}]%
        {wang2007jeopardy}
\bibfield{author}{\bibinfo{person}{Mengqiu Wang}, \bibinfo{person}{Noah~A
  Smith}, {and} \bibinfo{person}{Teruko Mitamura}.}
  \bibinfo{year}{2007}\natexlab{}.
\newblock \showarticletitle{What is the Jeopardy Model? A Quasi-Synchronous
  Grammar for QA.}. In \bibinfo{booktitle}{\emph{Proc. of EMNLP-CoNLL}},
  Vol.~\bibinfo{volume}{7}. \bibinfo{pages}{22--32}.
\newblock


\bibitem[\protect\citeauthoryear{Yang, Yih, and Meek}{Yang
  et~al\mbox{.}}{2015}]%
        {yang2015wikiqa}
\bibfield{author}{\bibinfo{person}{Yi Yang}, \bibinfo{person}{Wen-tau Yih},
  {and} \bibinfo{person}{Christopher Meek}.} \bibinfo{year}{2015}\natexlab{}.
\newblock \showarticletitle{WikiQA: A Challenge Dataset for Open-Domain
  Question Answering.}. In \bibinfo{booktitle}{\emph{Proc. of EMNLP}}.
  Citeseer, \bibinfo{pages}{2013--2018}.
\newblock


\bibitem[\protect\citeauthoryear{Yin, Sch{\"u}tze, Xiang, and Zhou}{Yin
  et~al\mbox{.}}{2015}]%
        {yin2015abcnn}
\bibfield{author}{\bibinfo{person}{Wenpeng Yin}, \bibinfo{person}{Hinrich
  Sch{\"u}tze}, \bibinfo{person}{Bing Xiang}, {and} \bibinfo{person}{Bowen
  Zhou}.} \bibinfo{year}{2015}\natexlab{}.
\newblock \showarticletitle{Abcnn: Attention-based convolutional neural network
  for modeling sentence pairs}.
\newblock \bibinfo{journal}{\emph{arXiv preprint arXiv:1512.05193}}
  (\bibinfo{year}{2015}).
\newblock


\bibitem[\protect\citeauthoryear{Yu, Hermann, Blunsom, and Pulman}{Yu
  et~al\mbox{.}}{2014}]%
        {yu2014deep}
\bibfield{author}{\bibinfo{person}{Lei Yu}, \bibinfo{person}{Karl~Moritz
  Hermann}, \bibinfo{person}{Phil Blunsom}, {and} \bibinfo{person}{Stephen
  Pulman}.} \bibinfo{year}{2014}\natexlab{}.
\newblock \showarticletitle{Deep learning for answer sentence selection}.
\newblock \bibinfo{journal}{\emph{arXiv preprint arXiv:1412.1632}}
  (\bibinfo{year}{2014}).
\newblock


\bibitem[\protect\citeauthoryear{Zhai}{Zhai}{2008}]%
        {DBLP:series/synthesis/2008Zhai}
\bibfield{author}{\bibinfo{person}{ChengXiang Zhai}.}
  \bibinfo{year}{2008}\natexlab{}.
\newblock \bibinfo{booktitle}{\emph{Statistical Language Models for Information
  Retrieval}}.
\newblock \bibinfo{publisher}{Morgan {\&} Claypool Publishers}.
\newblock


\bibitem[\protect\citeauthoryear{Zhang, Niu, Su, Wang, Ma, and Song}{Zhang
  et~al\mbox{.}}{2018}]%
        {aaaiZhang18}
\bibfield{author}{\bibinfo{person}{Peng Zhang}, \bibinfo{person}{Jiabin Niu},
  \bibinfo{person}{Zhan Su}, \bibinfo{person}{Benyou Wang},
  \bibinfo{person}{Liqun Ma}, {and} \bibinfo{person}{Dawei Song}.}
  \bibinfo{year}{2018}\natexlab{}.
\newblock \showarticletitle{End-to-End Quantum-like Language Models with
  Application to Question Answering}. In \bibinfo{booktitle}{\emph{Proc. of
  AAAI}}. \bibinfo{pages}{5666--5673}.
\newblock


\bibitem[\protect\citeauthoryear{Zhao, Zhang, Song, and Hou}{Zhao
  et~al\mbox{.}}{2011}]%
        {zhao2011novel}
\bibfield{author}{\bibinfo{person}{Xiaozhao Zhao}, \bibinfo{person}{Peng
  Zhang}, \bibinfo{person}{Dawei Song}, {and} \bibinfo{person}{Yuexian Hou}.}
  \bibinfo{year}{2011}\natexlab{}.
\newblock \showarticletitle{A novel re-ranking approach inspired by quantum
  measurement}. In \bibinfo{booktitle}{\emph{Proc. of ECIR}}.
  \bibinfo{pages}{721--724}.
\newblock


\bibitem[\protect\citeauthoryear{Zuccon and Azzopardi}{Zuccon and
  Azzopardi}{2010}]%
        {DBLP:conf/ecir/ZucconA10}
\bibfield{author}{\bibinfo{person}{Guido Zuccon} {and} \bibinfo{person}{Leif
  Azzopardi}.} \bibinfo{year}{2010}\natexlab{}.
\newblock \showarticletitle{Using the Quantum Probability Ranking Principle to
  Rank Interdependent Documents}. In \bibinfo{booktitle}{\emph{Proc. of ECIR}}.
  \bibinfo{pages}{357--369}.
\newblock


\end{thebibliography}

\end{document}